\documentclass[letterpaper, 10 pt, conference]{IEEEtran}
\IEEEoverridecommandlockouts
\usepackage[linkcolor=black,colorlinks=true]{hyperref}
\usepackage{amsmath,amssymb, graphicx, epstopdf,cite,enumerate,booktabs, setspace, float,stfloats,bm, multirow, lscape}
\usepackage{flushend}
\usepackage[normalem]{ulem}
\usepackage{diagbox}
\usepackage{float}
\usepackage{caption}
\usepackage{subcaption}
\usepackage{graphbox}
\usepackage{soul, color, xcolor}
\usepackage[bottom=1.52cm,top=2.01cm,left=1.69cm,right=1.69cm]{geometry}
\usepackage[linesnumbered,ruled]{algorithm2e}

\allowdisplaybreaks[4]

\title{Edge Accelerated Robot Navigation With Collaborative Motion Planning
}
\author{Guoliang Li$^{*}$, Ruihua Han$^{*}$, Shuai Wang, \emph{Senior Member, IEEE}, Fei Gao, \emph{Member, IEEE}, \\Yonina C. Eldar, \emph{Fellow, IEEE}, and Chengzhong Xu, \emph{Fellow, IEEE}
\thanks{Manuscript received November 9, 2023; revised April 15, 2024; accepted June 1, 2024. This work was supported by the Science and Technology Development Fund of Macao S.A.R (FDCT) (No. 0123/2022/AFJ and No. 0081/2022/A2), the National Natural Science Foundation of China (No. 62371444), the Guangdong Basic and Applied Basic Research Project (No. 2021B1515120067), and the Direct Drive Tech Cooperation Project. \emph{($\ast$ indicates equal contribution.)} \emph{(Corresponding authors: Shuai Wang and Chengzhong Xu.)}} 
\thanks{Guoliang Li and Chengzhong Xu are with the State Key Laboratory of Internet of
Things for Smart City (SKL-IOTSC), Department of Computer and Information Science, University of Macau, Macau 999078, China (e-mail: li.guoliang@connect.um.edu.mo; czxu@um.edu.mo).}
\thanks{Ruihua Han is with the Department of Computer Science, The University of Hong Kong,
Hong Kong (e-mail: hanrh@connect.hku.hk).}
\thanks{Shuai Wang is with the Shenzhen Institute of Advanced Technology, Chinese Academy of Sciences, Shenzhen 518055, China (e-mail: s.wang@siat.ac.cn).}
\thanks{Fei Gao is with the State Key Laboratory of Industrial Control Technology
and the Huzhou Institute, Zhejiang University, Zhejiang 310027, China
(e-mail: fgaoaa@zju.edu.cn).}
\thanks{Yonina C. Eldar is with the Faculty of Mathematics and Computer
Science, Weizmann Institute of Science, Rehovot 7610001, Israel (e-mail: yonina.eldar@weizmann.ac.il).}
\thanks{The code will be released at \url{https://github.com/GuoliangLI1998/EARN}.}
}

\begin{document}
\maketitle
\setlength{\textfloatsep}{3pt}

\begin{abstract}
Low-cost distributed robots suffer from limited onboard computing power, resulting in excessive computation time when navigating in cluttered environments.
This paper presents Edge Accelerated Robot Navigation (EARN), to achieve real-time collision avoidance by adopting collaborative motion planning (CMP). 
As such, each robot can dynamically switch between a conservative motion planner executed locally to guarantee safety (e.g., path-following) and an aggressive motion planner executed non-locally to guarantee efficiency (e.g., overtaking). 
In contrast to existing motion planning approaches that ignore the interdependency between low-level motion planning and high-level resource allocation, EARN adopts model predictive switching (MPS) that maximizes the expected switching gain with respect to robot states and actions under computation and communication resource constraints.
The MPS problem is solved by a tightly-coupled decision making and motion planning framework based on bilevel mixed-integer nonlinear programming and penalty dual decomposition. We validate the performance of EARN in indoor simulation, outdoor simulation, and real-world environments. Experiments show that EARN achieves significantly smaller navigation time and higher success rates than state-of-the-art navigation approaches.
\end{abstract}
\begin{IEEEkeywords}
Collaborative computing, distributed robotics, motion planning, model predictive switching.
\end{IEEEkeywords}
\section{Introduction}\label{section1}

Navigation is a fundamental task for mobile robots, which determines a sequence of control commands to move the robot safely from its current state to a target state. 
The navigation time and success rate depend on how fast the robot can compute a trajectory. {The computation time is proportional to the number of obstacles (spatial) and the length of prediction horizon (temporal) \cite{xia2022trajectory}.} In cluttered environments with numerous obstacles, the shape of obstacles should also be considered, and the computation time is further multiplied by the number of surfaces of each obstacle \cite{zhang2020optimization}. Therefore, navigation in cluttered environments is challenging for low-cost robots.

\begin{figure}[!t]
\centering
      \includegraphics[width=0.46\textwidth]{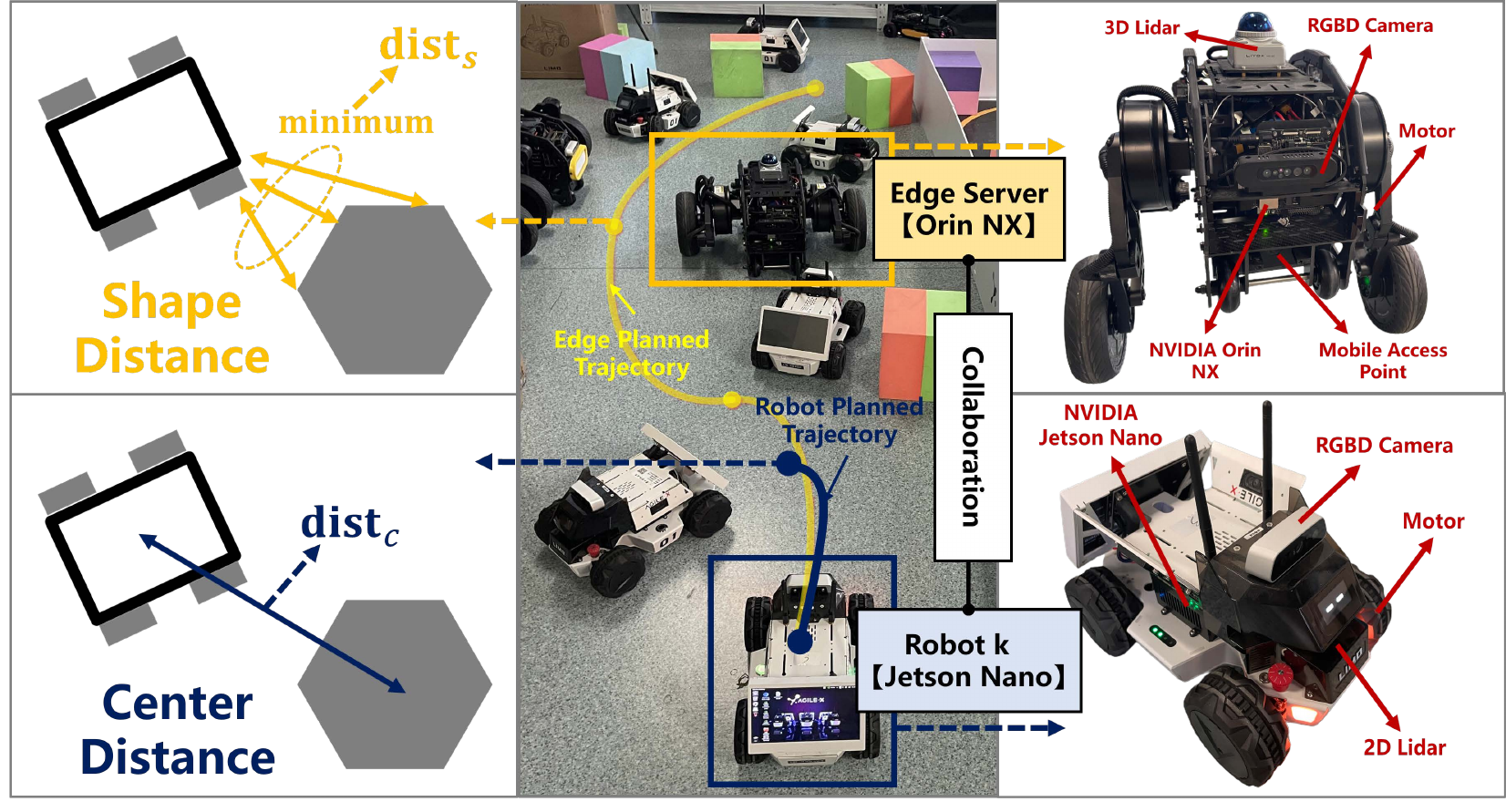}
      \caption{{Low-cost robots execute shape distance collision avoidance assisted by a proximal edge server based on EARN.}}
      \label{fig:eyecatcher}
\end{figure}

Currently, most existing approaches reduce the computation time from an algorithm design perspective, e.g., using heuristics \cite{wang2022extremum}, approximations \cite{han2023efficient}, parallelizations \cite{rey2018fully,wang2018parallel,han2022rda,zhenmin2023}, or learning \cite{xiao2020multimodal,pan2020imitation,kou2023iros} techniques. 
This paper proposes Edge Accelerated Robot Navigation (EARN), which accelerates the robot navigation from a collaborative computing for motion planning perspective \cite{sr2019,RILaaS,hayat2021edge}.
As shown in Fig.~\ref{fig:eyecatcher}, with only onboard computer (i.e., Jetson Nano), the robot can only adopt center distance collision avoidance, which generates a short trajectory (marked in blue) blocked by the narrow gap. In contrast, by leveraging EARN, the robot can opportunistically 
execute shape distance collision avoidance by accessing a proximal edge server  (i.e., Orin NX), which generates a long trajectory (marked in yellow) reaching the goal.

However, design and implementation of EARN is non-trivial. First, it involves periodic data exchange between robots and servers. Therefore, the server should have low-latency access to the robot; otherwise, communication delays may lead to collisions. Second, different robots may compete for a common computing server, and the computation time increases quickly as the number of robots increases. Third, in case of no proximal server, the robots should be able to navigate individually with only onboard computing resources. 
These observations imply that EARN needs to maximize the navigation efficiency under practical resource constraints, for which the existing local \cite{rey2018fully,wang2018parallel,han2022rda,zhenmin2023} or edge \cite{sr2019,RILaaS,hayat2021edge} planning methods become inefficient and unsafe, as they ignore the interdependence between low-level motion planning (e.g., robot states and actions) and high-level decision-making (e.g., planner switching and resource management).

To address all the above challenges, this paper proposes collaborative motion planning (CMP) and model predictive switching (MPS) to realize tightly-coupled decision making and motion planning (T-DMMP). The MPS maximizes the \emph{expected switching gain} under computation and communication resource constraints, which is in contrast to existing model predictive control (MPC) where 
robot states and actions are the only considerations.
In particular, for high-level resource management, 
we propose a bilevel mixed-integer nonlinear programming (B-MINLP) algorithm for decision-making, which can automatically identify switching-beneficial robots while orchestrating computation and communication resources.
For low-level motion planning, we incorporate the high-level decision variables into a set of conditional collision avoidance constraints, and propose a penalty dual decomposition (PDD) algorithm, which computes collision-free trajectories in parallel with convergence guarantee. We implement our methods by Robot Operation System (ROS) and validate them in the high-fidelity Car-Learning-to-Act (CARLA) simulation. Experiments confirm the superiority of the proposed scheme compared with various benchmarks in both indoor and outdoor scenarios.
We also implement EARN in a multi-robot testbed, where real-world experiments are conducted to demonstrate the practical applicability of EARN.

The contribution of this paper is summarized as follows:
\begin{itemize}
    \item Propose EARN, which enables T-DMMP in dynamic environments\footnote{{States (including poses and velocities) of the obstacles and resources  (including communication and computation) of the system are both dynamic.}} under resource constraints based on MPS.
    \item Propose B-MINLP and PDD algorithms, which ensure smooth planner switching and real-time motion planning.
    \item Implementation of EARN in both simulation and real-world environments. \item Evaluations of the performance gain brought by EARN compared with extensive benchmarks.
\end{itemize}

The remainder of the paper is organized as follows. Section \ref{section2} reviews the related work. Section \ref{section3} presents the architecture and mechanism of the proposed EARN. Section \ref{section4} presents the core optimization algorithms for EARN. Simulations and experiments are demonstrated and analyzed in Section \ref{section5}. 
Finally, conclusion is drawn in Section \ref{section6}.

\section{Related Work}\label{section2} 

\textbf{Motion Planning} is a challenge when navigating low-cost robots, due to contradiction between the stringent timeliness constraint and the limited onboard computing capability.
Conventional heuristic methods (e.g., path following \cite{wang2022extremum}, spatial cognition \cite{liu2022robotic}) are over-conservative, and the robot may get stuck in cluttered environments. 
Emerging optimization techniques can overcome this issue by explicitly formulating collision avoidance as distance constraints.
Model predictive control (MPC) \cite{zhang2020optimization,wang2022extremum, xia2022trajectory,rey2018fully,wang2018parallel,han2022rda,zhenmin2023} is the most widely used optimization algorithm, which leverages constrained optimization for generating high-performance collision-free trajectories in complex scenarios. Nonetheless, MPC could be time-consuming when the number of obstacles is large.  

\textbf{Optimization Based Collision Avoidance} can be accelerated in two ways. First, imitation learning methods \cite{xiao2020multimodal,pan2020imitation, kou2023iros} can learn from optimization solvers' demonstrations using deep neural networks.
As such, iterative optimization is transformed into a feed-forward procedure that can generate actions in milliseconds.
However, learning-based methods may break down if the target scenario contains examples outside the distribution of the training dataset \cite{kou2023iros}. Second, the computation time can be reduced by parallel optimizations \cite{rey2018fully,wang2018parallel,han2022rda,zhenmin2023}.
For instance, the alternating direction method of multipliers (ADMM) has been adopted in multi-robot navigation systems \cite{zhenmin2023}, which decomposes
a large centralized problem into small subproblems that are solved in parallel for each robot.
By applying parallel computation to obstacle avoidance, ADMM has been shown to accelerate autonomous navigation in cluttered environments \cite{han2022rda}. 
However, \emph{ADMM may diverge when solving nonconvex problems \cite{pdd}. 
Unfortunately, motion planning problems are nonconvex due to the nonlinear vehicle dynamics and irregular obstacle shapes \cite{zhang2020optimization,han2022rda}.
Here, we develop a PDD planner that converges to a Karush-Kuhn-Tucker (KKT) solution and overcomes the occasional failures of ADMM. }

\textbf{Cloud and Edge Robotics} are emerging paradigms to accelerate robot navigation \cite{sr2019} and learning \cite{liu2020federated,liu2021peer}. 
The idea is to allow robots to access proximal computing resources.
For example, robot inference and learning applications, such as object recognition and grasp planning, can be offloaded to cloud, edge, and fog as a service \cite{RILaaS}.
Edge-assisted autonomous driving was investigated in \cite{Syeda}, which offloads the heavy tasks from low-cost vehicular computers to powerful edge servers.
However, this type of approach would introduce additional communication latency \cite{huang2022edge}.
To reduce the communication latency, a partial offloading scheme was proposed for vision-based robot navigation \cite{hayat2021edge}.
A priority-aware robot scheduler was proposed to scale up collaborative visual simultaneous localization and mapping (SLAM) service in edge offloading settings \cite{nsdi2022}.
A cloud robotics platform FogROS2 was proposed to effectively connect robot systems across different physical locations, networks, and data distribution services \cite{FogROS2}. In recent DARPA SubT challenge \cite{morrell2022nebula}, it is found that for multi-robot motion planning, it is necessary to develop a higher-level ``mission planner'' that assigns different tasks to different robots according to their states and capabilities.
Nonetheless,  \emph{current cloud and edge robotics schemes ignore the interdependence between the low-level motion planning (e.g., robot states and actions) and the high-level decision-making (e.g., planner switching and resource management).
In contrast, EARN models such interdependence explicitly through MPS and achieves the T-DMMP feature by two optimization algorithms.}
Note that our method belongs to vertical collaboration that split computing loads between the server and the robot, which differs from conventional horizontal collaboration, e.g., cooperative motion planning \cite{rey2018fully}.

\begin{figure}[t]
  \centering
    \includegraphics[width=0.48\textwidth]{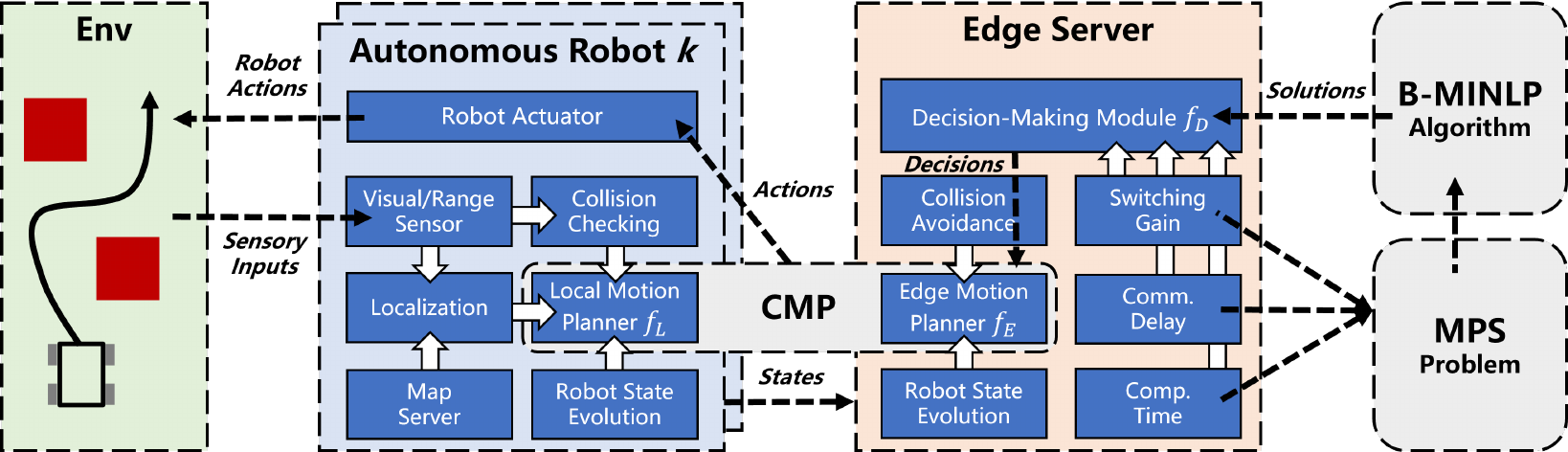}
    \caption{Architecture of EARN.}
  \label{fig:EARN_Arch}
\end{figure}

\section[]{Tightly Coupled Decision-Making and \\Motion Planning}\label{section3} 

The architecture of EARN is shown in Fig.~\ref{fig:EARN_Arch}, which adopts a decision-making module $f_{D}$ to execute planner switching between a low-complexity local motion planner $f_{L}(\cdot)$ deployed at $K$ robots and a high-performance edge motion planner $f_E(\cdot)$ deployed at the edge server.
Decision-making operates at a low frequency (e.g., $1$\,Hz) and motion planning operates at a high frequency (e.g., $10\sim100$\,Hz). 
In the following subsections, we first present the CMP formulation that determines $f_{L}$ and $f_E$. 
Then we present the MPS formulation that determines $f_{D}$.

\subsection{Collaborative Motion Planning}

At the $t$-th time slot, the $k$-th robot (with $1\leq k\leq K$) estimates its current state 
$\mathbf{s}_{k,t}=(x_{k,t},y_{k,t},\theta_{k,t})$ and obstacle states $\{\mathbf{o}_{1,t},...,\mathbf{o}_{M,t}\}$ with $\mathbf{o}_{m,t}=(a_{m,t},b_{m,t},\phi_{m,t})$ (with $1\leq m\leq M$), where $(x_{k,t},y_{k,t})$ and $\theta_{k,t}$ are positions and orientations of the $k$-th robot, and $(a_{m,t},b_{m,t})$ and $\phi_{m,t}$ are positions and orientations of the $m$-th obstacle.
The edge server collects the state information from $K$ robots and executes the decision-making module $f_{D}$. The output of $f_{D}$ is a set of one-zero decision variables $\{\alpha_1,\cdots,\alpha_K\}\in\{0,1\}^K$, where $\alpha_k=1$ represents the $k$-th robot being selected for edge planning and $\alpha_k=0$ represents the $k$-th robot being selected for local planning. Given $\alpha_k$, the local planner or edge planner can generate control vectors by minimizing the distances between the robot's footprints $\{\mathbf{s}_{k,t}\}$ and the target waypoints $\{\mathbf{s}_{k,t}^\diamond\}$.
Denoting the current time as $t=0$, the state evolution model $\mathbf{s}_{k,t+1}=E_k(\mathbf{s}_{k,t},\mathbf{u}_{k,t})$ is adopted to predict the future trajectories $\{\mathbf{s}_{k,t}\}_{t=0}^H$, where $H$ is the length of prediction horizon and $E_k$ is determined by Ackermann kinetics:
\begin{equation}
E_k\left(\mathbf{s}_{k,t},\mathbf{u}_{k,t}\right) = \mathbf{A}_{k,t}{\mathbf{s}_{k,t}} + \mathbf{B}_{k,t}{\mathbf{u}_{k,t}} + \mathbf{c}_{k,t},~\forall k,t,
  \label{dynamics}
\end{equation}
where $(\mathbf{A}_{k,t}$, $\mathbf{B}_{k,t}$, $\mathbf{c}_{k,t})$ are coefficient matrices 
defined in equations (8)--(10) of \cite[Sec. III-B]{han2022rda}. Furthermore, based on the state evolution model, we compute the distance $\{{\bf{dist}}\left(\mathbf{s}_{k,t},\mathbf{o}_{m,t}\right)\}_{t=0}^H$ between the $k$-th robot and the $m$-th obstacle at all time slots within the horizon, where the states $\{\mathbf{o}_{m,t}\}_{t=1}^H$ are obtained from motion prediction. Consequently, the CMP problem for the $k$-th robot is formulated as
\begin{subequations} 
\vspace{-0.1in}
    \begin{align}
    &\min \limits_{\{\mathbf{s}_{k,t}, \mathbf{u}_{k,t}\}_{t=0}^H}~~ \sum^{H}_{t=0} 
    \left\|\mathbf{s}_{k,t}-\mathbf{s}_{k,t}^\diamond\right\|^2 \label{problemAa} \\
    &\text { s.t. }~~~ \mathbf{s}_{k,t+1}=
E_k(\mathbf{s}_{k,t},\mathbf{u}_{k,t}),~\forall t,\label{problemAb}\\
    &~~ ~~~ ~~~~ \mathbf{u}_{\min } \preceq \mathbf{u}_{k,t} \preceq \mathbf{u}_{\max },~\forall t,  \label{problemAc}
    \\
    &~~ ~~~ ~~~~ \mathbf{a}_{\min }  \preceq  {\mathbf{u}_{k,t+1}}
    -{\mathbf{u}_{k,t}}   \preceq \mathbf{a}_{\max },~\forall t, \label{problemAd}
    \\
    &~~ ~~~ ~~~~ {\Psi\left(\mathbf{s}_{k,t},\mathbf{o}_{m,t},d_{\text{safe}}|\alpha_k \right)\geq 0,~\forall m\in\mathcal{M}(k), t,} \label{problemAe}
    \end{align}
\end{subequations}
where $\Psi$ is a conditional collision avoidance function 
\begin{align}
&\Psi\left(\mathbf{s}_{k,t},\mathbf{o}_{m,t},d_{\text{safe}}|\alpha_k \right)=\alpha_k\left({\bf{dist}}\left(\mathbf{s}_{k,t},\mathbf{o}_{m,t}\right)-d_{\text{safe}}\right),
\label{conditional}
\end{align}
and $\mathcal{M}(k)$ is the set of obstacles within the local map of robot $k$ with its cardinality denoted as $|\mathcal{M}(k)|$.
Constants ${\mathbf{u}_{\min }}$ (${\mathbf{a}_{\min }}$) and ${\mathbf{u}_{\max }}$ (${\mathbf{a}_{\max }}$) are the minimum and maximum limits of the control (acceleration) vector, respectively. 

Now consider two cases. 
If $\alpha_k=1$, the collision avoidance constraints exist, and the associated optimal solution $\{\mathbf{s}^{[1]}_{k,t},\mathbf{u}^{[1]}_{k,t}\}_{t=0}^H$ to problem (2) leads to a \emph{proactive} collision avoidance policy. 
In this case, problem (2) is solved at the edge server and we define $f_E(\mathbf{s}_{k,t},\{\mathbf{o}_{m,t}\}_{m=1}^M)=\mathbf{u}^{[1]}_{k,t}$. 
On the other hand, if $\alpha_k=0$, the collision avoidance constraints are discarded, and the optimal $\{\mathbf{s}^{[0]}_{k,t},\mathbf{u}^{[0]}_{k,t}\}_{t=0}^H$ to problem (2) leads to a \emph{reactive} collision avoidance policy.
In this case, problem (2) is solved at the robot $k$ and we define $f_L(\mathbf{s}_{k,t},\{\mathbf{o}_{m,t}\}_{m=1}^M)=\mathbf{u}^{[0]}_{k,t}$ if the robot sensor detects no obstacles ahead within a certain braking distance $d_{\mathrm{B}}$ and $f_L(\mathbf{s}_{k,t},\{\mathbf{o}_{m,t}\}_{m=1}^M)=\mathbf{u}_{B}$ otherwise, where $\mathbf{u}_{B}$ denotes the braking action.

\subsection{Model Predictive Switching}

The decision making module aims to find the optimal decision variables $\{\alpha_k^*\}$ by maximizing the expected switching gain over all robots under the communication and computation resource constraints. 
First, the communication latency between the $k$-th robot and the edge server should be smaller than a certain threshold $D_{\mathrm{th}}$ if $\alpha_k=1$, i.e.,
$\alpha_kD_k(\mathbf{s}_{k,t})\leq D_{\mathrm{th}}$, where $D_k$ is a function of the robot state $\mathbf{s}_{k,t}$.
Second, the total computation time at the server should not exceed a certain threshold $C_{\mathrm{th}}$, i.e., 
$\sum_k\alpha_kC_k(H,|\mathcal{M}(k)|)\leq C_{\mathrm{th}}$, 
where $C_k$ is a function of the prediction time length $H$ and the number of obstacles $|\mathcal{M}(k)|$. 
Third, the expected switching gain is defined as the difference between the moving distance of planner $f_L$ and that of $f_E$ at time $H$, which is 
\begin{align}\label{G}
&G_k\left({\mathbf{s}_{k,t}},\{\mathbf{o}_{m,t}\},f_E,f_L\right)
\nonumber\\
&
=
{\bf{dist}}_c(\mathbf{s}^{{[1]}}_{k,H},\mathbf{s}_{k,0})-{\bf{dist}}_c(\mathbf{s}^{{[0]}}_{k,H},\mathbf{s}_{k,0}).
\end{align}
where 
\begin{align}{\bf{dist}}_c(\mathbf{s}^{{[1]}}_{k,H},\mathbf{s}_{k,0})=\sqrt{(x^{{[1]}}_{k,H}-x_{k,0})^2+(y^{{[1]}}_{k,H}-y_{k,0})^2}, \label{dist_c}
\end{align}
represents the center distance between the states $\mathbf{s}^{{[1]}}_{k,H}$ and $\mathbf{s}_{k,0}$. With the above models, the MPS problem is formulated as
\begin{subequations} 
    \begin{align}
    \max \limits_{\{\alpha_k\}_{k=1}^K}~~
    &\sum_{k=1}^K\alpha_kG_k\left({\mathbf{s}_{k,t}},\{\mathbf{o}_{m,t}\},f_E,f_L\right) \\
    \text{ s.t. }~~&
    \{\mathbf{s}^{{[\alpha_k]}}_{k,H}\}=\mathop{\arg\,\min \limits_{\{\mathbf{s}_{k,t}, \mathbf{u}_{k,t}\}_{t=0}^H}}
    \Big\{
    \sum^{H}_{t=0} \nonumber\\
    &
    \left\|\mathbf{s}_{k,t}-\mathbf{s}_{k,t}^\diamond\right\|^2:
    (\ref{problemAb})- (\ref{problemAe})\Big\},~\forall k,
    \label{problemBb}
    \\
    &\alpha_kD_k(\mathbf{s}_{k,t})\leq D_{\mathrm{th}},~\forall k,
    \label{problemBc}
    \\
    &\sum_{k=1}^K\alpha_kC_k(H,|\mathcal{M}(k)|)\leq C_{\mathrm{th}}, \label{problemBd}
    \\
    &\alpha_k\in\{0,1\},~\forall k. \label{problemBe}
    \end{align}
\end{subequations}
Denoting the optimal solution to MPS as $\{\alpha_k^*\}_{k=1}^K$, we set $f_D(\mathbf{s}_{k,t},\{\mathbf{o}_{m,t}\}_{m=1}^M,D_{\mathrm{th}},C_{\mathrm{th}})=\{\alpha_k^*\}_{k=1}^K$. 
It can be seen that MPS requires joint considerations of robot states, obstacle states, communications, and computations. Inappropriate decisions would lead to collisions due to local/edge computation timeout or communication timeout.

\begin{table}[t]
    \centering
    \caption{{Comparison of different schemes}}
        \scalebox{0.75}{
    \begin{tabular}{|c|c|c|c|c|}
        \hline
        \multirow{2}{*}{\diagbox{Metric}{Scheme}} & \multirow{2}{*}{EARN} & \multirow{2}{*}{Onboard Comp.} & \multirow{2}{*}{Enhanced Onboard Comp.}
        \\ &  &  & \\
        \hline
        Latency (ms) & 50 & 200 & 50 \\\hline
        Power (W) & 65 & 50 & 125   
        \\\hline
        Price (\$) & 1600 & 750 & 5000 
         \\\hline
    \end{tabular}
    \label{table_I}
        }
\end{table}

\subsection{Case Study}

To demonstrate the practicability of EARN, a case study is conducted for a 5-robot system. We consider two embedded computing chips, i.e., NVIDIA Jetson Nano and NVIDIA Orin NX, and two motion planners, i.e., PF (reactive) \cite{wang2022extremum} and RDA (proactive) \cite{han2022rda}.
The power and price of Jetson Nano are 10 Watts and 150\,\$. The power and price of Orin NX are 25 Watts and 1000\,\$. Based on our experimental data, 
a frequency of $50\,$Hz can be achieved for PF on both chips.
RDA can be executed at a frequency of $5\,$Hz on Jetson Nano, and $20\,$Hz on Orin NX, when the number of obstacles is $5$ \cite{han2022rda}.

We consider three schemes: 1) onboard computing, where all robots execute RDA on Jetson Nano; 2) enhanced onboard computing, where all robots execute RDA on Orin NX; 3) EARN, where 4 low-cost robots execute PF on Jetson Nano and 1 edge server execute RDA on Orin NX. 
The latency, power, and price of different schemes are illustrated in Table \ref{table_I}, and we have the following observations: 1) The average computation latency of EARN ($50$\,ms) is significantly smaller than that of the onboard computing ($200$\,ms); 2) EARN ($65$\,W) is more energy efficient than the enhanced onboard computing ($125$\,W); 3) The total system cost of EARN is significantly smaller than that of enhanced onboard computing (1600\,\$ versus 5000\,\$).

In summary, EARN is faster than onboard computing, and their time difference is the \textbf{acceleration gain}.
On the other hand, EARN is more energy-and-cost efficient than enhanced onboard computing, and their cost difference is the \textbf{computing sharing gain}.
The insight is that a robot is not always encountering challenging environments, and it is possible to use a single server to support multiple robots at different times.
The acceleration gain and computing sharing gain brought by EARN increase as the complexity of motion planners increases. This makes EARN suitable for large model empowered robot navigation at the edge.

\section{Algorithm Designs for Decision-Making and Motion Planning}\label{section4} 

\subsection{Decision-Making}

To solve the MPS problem, we need mathematical expressions of the communication latency $D_k$, the computation latency $C_k$, and the planner switching gain $G_k$. First, the function $D_k$ is determined by the signal attenuation $Q_k$ between the $k$-th robot and the server.
This $Q_k$ is a function of the robot position $\mathbf{s}_{k,t}$, which is known as radio map. 
It is not only related to the robot-edge distance, but also related to non-line-of-sight (nLOS) signal propagation (shadowing, reflection, diffraction, blockage) \cite{mobisys}. 
We adopt nvidia sionna, a widely used ray tracing package \cite{sionna}, to obtain the radio map (e.g., Fig.~\ref{fig:indoor_trajectory}b), where similar colors represent similar attenuation (and latency). 
As such, we can segment the radio map into $J$ regions by color and for the $j$-th region, we select representative positions and ping the communication latencies between the robot and the server. 
The measurements are adopted to estimate the latency distribution of the region. 
Consequently, we can write $D_k$ as 
\begin{align}
&D_k(\mathbf{s}_{k,t})
=
\sum_{j=1}^JT_l\,\mathbb{I}_{\mathcal{Z}_j}(\mathbf{s}_{k,t}), \label{commun_model}
\end{align}
where the indicator function $\mathbb{I}$ is given by 
\begin{equation}\label{zone}
\mathbb{I}_{\mathcal{Z}_j}(\mathbf{s}_{k,t})=
\left\{\begin{array}{ll}
1, & \mathrm{if}~\mathbf{s}_{k,t}\in\mathcal{Z}_j\\
0, &\mathrm{if}~\mathbf{s}_{k,t}\notin\mathcal{Z}_j
\end{array}\right.
,
\end{equation} 
and $\{\mathcal{Z}_1,\cdots,\mathcal{Z}_J\}$ are different areas of radio map, where $T_j$ represents the random communication latency of the $j$-th area with a known distribution.

Next, we determine the computing latency $C_k$. Since a common motion planner has a polynomial-time computational complexity, we can write $C_k$ as
\begin{align}
C_k=\gamma H|\mathcal{M}(k)|^p+\tau,
\label{comp_model}
\end{align}
where $\gamma$ and $\tau$ are hardware-dependent hyper-parameters estimated from historical experimental data. Moreover, $p$ is the order of complexity, which ranges from $1$ to $3.5$ and depends on the motion planner. 

Finally, we determine $G_k$, which is coupled with the motion planning algorithms $f_E$ and $f_L$. 
{In particular, if ${\bf{dist}}_c(\mathbf{s}_{k,t},\mathbf{o}_{m,t})>d_{\mathrm{B}}$, we have $G_k\leq0$, since the local planning problem is a relaxation of the edge planning problem and the solution of the former is at least no worse than the latter.}
On the other hand, if 
${\bf{dist}}_c(\mathbf{s}_{k,t},\mathbf{o}_{m,t})\leq d_{\mathrm{B}}$, 
then ${\bf{dist}}_c(\mathbf{s}^{{[0]}}_{k,H},{\mathbf{s}_{k,0}})=0$ as the braking signal is generated and the robot stops. Combining the above two cases, $G_k$ is equivalently written as 
\begin{align}
&G_k=I_k\,
{\bf{dist}}_c(\mathbf{s}^{{[1]}}_{k,H},{\mathbf{s}_{k,0}}),
\end{align}
where 
\begin{equation}\label{Ik}
I_k=
\left\{\begin{array}{ll}
1,
&\mathrm{if}~{\bf{dist}}_c(\mathbf{s}_{k,t},\mathbf{o}_{m,t})\leq d_{\mathrm{B}}, 
{\mathbf{o}_{m,t}}\in\mathcal{W}_k
\\
0, 
& \mathrm{otherwise}
\end{array}\right.
,
\end{equation}
and set $\mathcal{W}_k$ represents the global path formed by waypoints $\{\mathbf{s}_{k,t}^\diamond\}$.
Note that we have replaced $G_k\leq 0$ with $G_k=0$, which would not affect the solution of the problem.

Based on the above derivations of $D_k$, $C_k$, and $G_k$, the original MPS problem (6) is transformed into
\begin{subequations} 
    \begin{align}
    \max \limits_{\{\alpha_k\}_{k=1}^K}~~
    &\sum_{k=1}^K\alpha_kI_k
{\bf{dist}}_c(\mathbf{s}^{{[1]}}_{k,H},{\mathbf{s}_{k,0}})  \label{obj} \\
    \text{ s.t. }~~&
    \textsf{constraints }(\ref{problemBb}), (\ref{problemBe}),
    \label{problemCb}
    \\
    &\alpha_k\sum_{l=1}^LT_l\,\mathbb{I}_{\mathcal{Z}_l}(\mathbf{s}_{k,t})\leq D_{\mathrm{th}},~\forall k, \label{problemCc}\\
    &\sum_{k=1}^K\alpha_k\gamma H|\mathcal{M}(k)|^p+\tau\leq C_{\mathrm{th}}. \label{problemCd}
    \end{align}
\end{subequations}
This is a B-MINLP problem
due to the constraint \eqref{problemCb}.
A naive approach is to solve the inner problem \eqref{problemBb} for all robots and conduct an exhaustive search over $\{\alpha_k\}$.
However, the resultant complexity would be 
$\mathcal{O}(2^K)$, which cannot meet the frequency requirement of the decision-making module. To this end, we propose a low-complexity method summarized in Algorithm 1, which consists of three sequential steps: 1) Prune out impossible solutions for space reduction. Specifically, we conduct the feasibility check of \eqref{problemCc}--\eqref{problemCd} and switching gain check of $\{I_k\}$ for all the robots. This yields sets $\mathcal{A}$ (robots satisfying \eqref{problemCc}--\eqref{problemCd} and having positive values of $I_k$) and $\mathcal{I}$ (otherwise).
Any robot in $\mathcal{I}$ is pruned out, without the need for further motion planning; 2) Leverage parallel motion planning for fast trajectory computations; 3) Put $\{\alpha_k=0\}_{k\in\mathcal{I}}$ and 
$\{\mathbf{s}^{{[1]}}_{k,t}\}_{k\in\mathcal{A}}$ into problem (12), and solve the resultant problem using integer linear programming. The integer programming can be either solved by CVXPY, or accelerated by other penalty continuous relaxation approaches.

\begin{algorithm}[t]
  \caption{B-MINLP decision making}
  \label{decision-making algorithm}
  \textbf{Input}: robots' states
  $\{\mathbf{s}_{k,t}\}_{k=1}^K$, obstacles' states $\{\mathbf{o}_{m,t}\}_{m=1}^M$, communication latency threshold $D_{\mathrm{th}}$, computation latency threshold $C_{\mathrm{th}}$. \\
\textbf{Initialize}: $\mathcal{A}=\mathcal{I}=\emptyset$
  \\
  \For{robot $k=1, \cdots,K$}
  {
    \eIf{$\alpha_k=1$ satisfies \eqref{problemCc}--\eqref{problemCd} and $I_k=1$}
    {Set $\alpha_k=1$ and update $\mathcal{A}\leftarrow \{k\}\cup\mathcal{A}$}
    {Set $\alpha_k=0$ and update $\mathcal{I}\leftarrow \{k\}\cup\mathcal{I}$}
 }
   \For{robot $k\in\mathcal{A}$}
  {
Compute $\{{\mathbf{s}^{{[1]}}_{k,t}}\}$ in \eqref{problemBb} using PDD}
Solve (12) with integer linear programming.\\
\textbf{Output}: 
$f_D(\mathbf{s}_{k,t},\{\mathbf{o}_{m,t}\}_{m=1}^M,D_{\mathrm{th}},C_{\mathrm{th}})=\{\alpha_k^*\}_{k=1}^K$.
\end{algorithm}

\subsection{Motion Planning}

When $\alpha_k=0$, the local motion planner is adopted. {To ensure safety, the distances between the robot and other obstacles are first computed using the onboard sensor. If the robot sensor detects any obstacle ahead within a certain distance $d_{\mathrm{B}}$, then the braking action $\mathbf{u}_{B}$ is directly adopted without further computation (hence this also reduces computation complexity).} Otherwise, the following planning problem (which corresponds to \eqref{problemBb} with $\alpha_k=0$) is considered
\begin{subequations} 
    \begin{align}
    \min \limits_{\{\mathbf{s}_{k,t}, \mathbf{u}_{k,t}\}_{t=0}^H}~~ &\sum^{H}_{t=0} 
    \left\|\mathbf{s}_{k,t}-\mathbf{s}_{k,t}^\diamond\right\|^2 \\
    \text { s.t. }~~ &\textsf{constraints }(\ref{problemAb})-(\ref{problemAd}).
    \end{align}
\end{subequations}
The above problem can be independently solved at each robot as there is no coupling among different $k$. 
Combining the above two cases, the action of robot $k$ can be represented as\\
\begin{equation}\label{fA_actions}
\mathbf{u}^{{[0]}}_{k,t}=
\left\{\begin{array}{ll}
\mathbf{u}_{\mathrm{B}},
&\mathrm{if}~{\bf{dist}}_c(\mathbf{s}_{k,t},\mathbf{o}_{m,t})\leq d_{\mathrm{B}},
{\mathbf{o}_{m,t}}\in\mathcal{W}_k
\\
\mathbf{u}^{*}_{k,t}, 
& \mathrm{otherwise}
\end{array}\right.
,
\end{equation} 
where $\{\mathbf{s}^{*}_{k,t},\mathbf{u}^{*}_{k,t}\}$ is the optimal solution to (13).

When $\alpha_k=1$, the edge motion planner is adopted, which corresponds to \eqref{problemBb} with $\alpha_k=1$:
\begin{subequations} 
    \begin{align}
    \min \limits_{\{\mathbf{s}_{k,t}, \mathbf{u}_{k,t}\}_{t=0}^H}~~ &\sum^{H}_{t=0} 
    \left\|\mathbf{s}_{k,t}-\mathbf{s}_{k,t}^\diamond\right\|^2 \\
    \text { s.t. }~~ &\textsf{constraints }(\ref{problemAb})-(\ref{problemAe}).
    \end{align}
\end{subequations}
To solve the above problem, we need an explicit form of ${\bf{dist}}\left(\mathbf{s}_{k,t},\mathbf{o}_{m,t}\right)$ in constraint \eqref{problemAe}. Conventional approaches \cite{jasontits} adopt center distance $
{\bf{dist}}\left(\mathbf{s}_{k,t},\mathbf{o}_{m,t}\right)=
{\bf{dist}}_c\left(\mathbf{s}_{k,t},\mathbf{o}_{m,t}\right)$, where each obstacle is modeled as a point. However, center distance model is not applicable to cluttered environments.
Emerging approaches \cite{zhang2020optimization,han2022rda} adopt shape distance 
$
{\bf{dist}}\left(\mathbf{s}_{k,t},\mathbf{o}_{m,t}\right)=
{\bf{dist}}_s({\mathbb{G}_{k,t}},{\mathbb{O}_{m,t}})$, where the $k$-th robot is modeled as a set $\mathbb{G}_{k,t}$ and the $m$-th obstacle is modeled as a set ${\mathbb{O}_{m,t}}$. 
However, function ${\bf{dist}}_s$ is an optimization problem itself, and needs to compute the minimum distance between any two points within $\mathbb{G}_{k,t}$ and 
$\mathbb{O}_{m,t}$.
This motivates us to develop a PDD motion planner for fast parallel optimization under shape distance model ${\bf{dist}}_s$. 
Note that PDD is different from RDA \cite{han2022rda}, as PDD incorporates the online calibration of penalty parameters into RDA, leading to convergence guaranteed robot navigation.

{To begin with, we define the shape models ${\mathbb{G}_{k,t}}$ and ${\mathbb{O}_{m,t}}$ for robot and obstacle \cite{han2022rda}.
We omit the subscript of robot index $k$ for simplicity, and set $\mathbb{G}_{t}$, which represents the geometric region occupied by the robot, is related to its state $\mathbf{s}_{t}$ and its shape $\mathbb{Z}$:
\begin{align}
  \mathbb{G}_{t}(\mathbf{s}_{t},
  \mathbb{Z}) &= \left\{\mathbf{z}\in \mathbb{Z}|\mathbf{R}(\mathbf{s}_{t})\mathbf{z} + \mathbf{p}(\mathbf{s}_{t})\right\},
\end{align}
where $\mathbf{R}(\mathbf{s}_{t}) \in {\mathbb{R}^{{3} \times {3}}}$ is the rotation matrix related to $\theta_{t}$ and $\mathbf{p}(\mathbf{s}_{t})$ is the translation vector related to $\left(x_{t},{y_{t}}\right)$. 
The set ${\mathbb{Z}}= \{ \mathbf{z}\in {\mathbb{R}^{3}}| \mathbf{G}\mathbf{z}\preceq 
     {\mathbf{g}}\}
$ represents the robot shape, where $\mathbf{G} \in {\mathbb{R}^{l \times 3}}$ and $\mathbf{g} \in {\mathbb{R}^{l}}$ ($l$ is the number of surfaces for the robot). Similarly, the $m$-th obstacle can be represented by 
$\mathbb{O}_{m,t}= \{ \mathbf{z}\in {\mathbb{R}^{3}}| {\mathbf{H}_{m,t}}\mathbf{z}\preceq 
     {\mathbf{h}_{m,t}}\}
$.
Consequently, constraint (\ref{problemAe}) is equivalently written as 
\begin{equation}
{\bf{dist}}_s({\mathbb{G}_{t}},{\mathbb{O}_{m,t}}) \ge {d_{\mathrm{safe}}},~\forall m\in\mathcal{M}(k),t.
\label{cac}
\end{equation}
Function ${\bf{dist}}_s$ is not analytical, but can be equivalently transformed into its dual form \cite{zhang2020optimization,han2022rda}
\begin{equation}
\begin{array}{l}
    {\left\| {{{\mathbf{H}_{m,t}}^T}\bm{\lambda}_{m,t} } \right\|} \leq 1,\\
    \bm{\lambda}_{m,t} \succeq \mathbf{0}, ~ \bm{\mu}_{m,t} { \succeq}\mathbf{0}, ~ z_{m,t}\geq0,\\
    {\bm{\mu}_{m,t}^T}{\mathbf{G}}
    +{\bm{\lambda}_{m,t} ^T}{\mathbf{H}_{m,t}}\mathbf{R}(\mathbf{s}_{t})
    =0,\\
    \bm{\lambda}_{m,t}^T{\mathbf{H}_{m,t}}\mathbf{p}(\mathbf{s}_{t}) -{\bm{\lambda}_{m,t}^T}{\mathbf{h}_{m,t}} \\- {\bm{\mu}_{m,t}^T}{\mathbf{g}} - z_{m,t} = {d}_{\mathrm{safe}},
\end{array}
\vspace{-0.05in}
\label{dual1}
\end{equation}
where $\bm{\lambda}_{m,t} \in {\mathbb{R}^{l_{m}}}$, $\bm{\mu}_{m,t}\in {\mathbb{R}^{l}}$ and $z_{m,t}\in \mathbb{R}$ are the dual variables representing our \emph{attentions} on different surfaces of robots and obstacles. Putting \eqref{dual1} into problem (15), the resultant problem is
\begin{subequations} 
    \begin{align}
    \min \limits_{\substack{\{\mathbf{s}_{t}, \mathbf{u}_{t}\}\\\{\bm{\lambda}_{m,t},\bm{\mu}_{m,t}, z_{m,t}\}}} &\sum^{H}_{t=0} 
    \left\|\mathbf{s}_{t}-\mathbf{s}_{t}^\diamond\right\|^2 \\
    \text { s.t. }~~ &\textsf{constraints }(\ref{problemAb})-(\ref{problemAd}), \\
    &\bm{\lambda}_{m,t} \succeq \mathbf{0}, \bm{\mu}_{m,t} { \succeq}\mathbf{0}, ~\forall m, t,\label{problem18a}\\
    &{\left\| {{{\mathbf{H}_{m,t}}^T}\bm{\lambda}_{m,t} } \right\|} \leq 1, z_{m,t}\geq 0, ~\forall m, t,\label{problem18b}\\
    &U_{m,t}\left(\mathbf{s}_t,\bm{\mu}_{m,t},\bm{\lambda}_{m,t}\right)=0, ~\forall m, t,\label{problem18c}\\
    &V_{m,t}\left(\mathbf{s}_t,\bm{\mu}_{m,t},\bm{\lambda}_{m,t},z_{m,t}\right)=0, ~\forall m, t,\label{problem18d}
    \end{align}
\end{subequations}
where we have defined 
\begin{align}
    U_{m,t}\left(\mathbf{s}_t,\bm{\mu}_{m,t},\bm{\lambda}_{m,t}\right)=
    {\bm{\mu}_{m,t}^T}{\mathbf{G}}
    +{\bm{\lambda}_{m,t} ^T}{\mathbf{H}_{m,t}}\mathbf{R}(\mathbf{s}_{t}),
\end{align}
\begin{align}
    V_{m,t}\left(\mathbf{s}_t,\bm{\mu}_{m,t},\bm{\lambda}_{m,t},z_{m,t}\right)=&\bm{\lambda}_{m,t}^T{\mathbf{H}_{m,t}}\mathbf{p}(\mathbf{s}_{t}) -{\bm{\lambda}_{m,t}^T}{\mathbf{h}_{m,t}}-\nonumber\\& {\bm{\mu}_{m,t}^T}{\mathbf{g}} - z_{m,t}-{d}_{\mathrm{safe}},
\end{align}
Now, instead of directly solving the dual problem using optimization software \cite{zhang2020optimization} or ADMM \cite{han2022rda}, we propose a PDD method (summarized in Algorithm 2) that constructs the following augmented Lagrangian
\begin{align}
    &\mathcal{L}
= \sum^{H}_{t=0} 
    \left\|\mathbf{s}_{t}-\mathbf{s}_{t}^\diamond\right\|^2
    +\frac{\rho}{2} \sum_{t=0}^{H} \sum_{m=1}^{M}\Big\|
    U_{m,t}\left(\mathbf{s}_t,\bm{\mu}_{m,t},\bm{\lambda}_{m,t}\right)
    \nonumber\\&+\bm{\zeta}_{m,t}\Big\|^{2} 
    +\frac{\rho}{2} \sum_{t=0}^{H} \sum_{m=1}^{M}\Big(V_{m,t}\left(\mathbf{s}_t,\bm{\mu}_{m,t},\bm{\lambda}_{m,t},z_{m,t}\right)
    +\xi_{m,t}\Big)^{2},
    \nonumber
\end{align}
where $\{\bm{\zeta}_{m,t},\xi_{m,t}\}$ are slack variables, $\rho$ is the penalty parameter. Then the augmented Lagrangian is minimized
\begin{align}
&
\mathop{\mathrm{min}}_{
\substack{
\{\mathbf{s}_{t},
\mathbf{u}_{t}\}\in\mathcal{X},
\\
\{(\bm{\lambda}_{m,t},\bm{\mu}_{m,t}, z_{m,t})\in\mathcal{Y}_{m,t}\}}} \mathcal{L}\Big(
\{\mathbf{s}_{t},{\mathbf{u}_{t}}\},
\nonumber\\
&~~~~~~~
\{\bm{\lambda}_{m,t},\bm{\mu}_{m,t}, z_{m,t}\};
\rho,\{\bm{\zeta}_{m,t},\xi_{m,t}\}
\Big),
\label{AL}
\end{align}
where $\mathcal{X}$ is the set for constraints 
\eqref{problemAb}--\eqref{problemAd},
$\mathcal{Y}$ is the set for constraint \eqref{problem18a}--\eqref{problem18b}.
To solve problem \eqref{AL}, we first optimize the robot states and actions with all other variables fixed via CVXPY as shown in line $4$ of Algorithm 2. 
Next, we optimize the collision attentions with all other variables fixed via CVXPY as shown in line $5$ of Algorithm 2. Lastly, from lines $6$ to $10$, we either execute dual update or penalty update based on the residual function:
\begin{align}
&\Phi\left(\{\bm{\lambda}_{m,t},\bm{\mu}_{m,t}, z_{m,t}\}\right)\nonumber\\
=&\max\Big\{
\max\limits_{m,t}~\Big\|U_{m,t}\left(\mathbf{s}_t,\bm{\mu}_{m,t},\bm{\lambda}_{m,t}\right)\Big\|_\infty,\nonumber\\
&\max\limits_{m,t}~\Big|V_{m,t}\left(\mathbf{s}_t,\bm{\mu}_{m,t},\bm{\lambda}_{m,t},z_{m,t}\right)\Big|
\Big\}.\nonumber
\end{align}
For dual update, we compute $\{\bm{\zeta}_{m,t}\},\{\xi_{m,t}\}$ using first-order method as line 7 in Algorithm 2.
For penalty update, we scale the penalty parameter as line 9 in Algorithm 2 with $\beta\geq1$ being an increasing factor.
According to \cite[Theorem 3.1]{pdd}, Algorithm 2 is guaranteed to converge to a KKT solution to problem \eqref{AL}. The complexity of PDD is given by $\mathsf{Comp}_k=\mathcal{O}((5H)^{3.5}+H\sum_{m\in\mathcal{M}(k)}(l_m+l_k)^{3.5}+H|\mathcal{M}(k)|(l_m+l_k))$.

\begin{algorithm}[!t]
  \caption{PDD motion planner}
  \label{pdd algorithm}
  \textbf{Input}: robot's state
  $\{\mathbf{s}_{t}\}$, obstacles' states $\{\mathbf{o}_{m}\}_{m=1}^M$ \\
  \textbf{Initialize}: robot's action $\{\mathbf{u}_{t}\}$, attentions $\{\bm{\lambda}_{m,t},\bm{\mu}_{m,t}, z_{m,t}\}$, and parameters $\{\beta,\rho,\eta_{iter}\}$
  \\
  \For{iteration $iter=1, 2,\cdots$}
  {$\{\mathbf{s}_{t}, \mathbf{u}_{t}\} \leftarrow \mathop{\mathrm{argmin}}_{\{\mathbf{s}_{t}, \mathbf{u}_{t}\} \in\mathcal{X}} \mathcal{L}$ \\ $(\bm{\lambda}_{m,t},\bm{\mu}_{m,t}, z_{m,t})\leftarrow \mathop{\mathrm{argmin}}_{ (\bm{\lambda}_{m,t},\bm{\mu}_{m,t}, z_{m,t})\in\mathcal{Y}_{m,t}}\mathcal{L},~\forall m,t$ \\
  \eIf{$\Phi\left(\{\bm{\lambda}_{m,t},\bm{\mu}_{m,t}, z_{m,t}\}\right)\leq \eta_{iter}$}{Update $\{\bm{\zeta}_{m,t},\xi_{m,t}\}$ using $\bm{\zeta}_{m,t}+U_{m,t}\left(\mathbf{s}_t,\bm{\mu}_{m,t},\bm{\lambda}_{m,t}\right),$ $~\xi_{m,t}+V_{m,t}\left(\mathbf{s}_t,\bm{\mu}_{m,t},\bm{\lambda}_{m,t},z_{m,t}\right)$}
  {Update $\rho\leftarrow \beta\rho$}}
  \textbf{Output}: $f_E(\mathbf{s}_{t},\{\mathbf{o}_{m,t}\}_{m=1}^M)=\mathbf{u}_{t}$.
\end{algorithm}

{\emph{Remark}: For practical implementation, a static safety distance $d_{\text{safe}}$ may not be suitable for all the time slots $t\in[0,H]$ since the state evolution starting from a feasible state may end up at a state that the safety constraint \eqref{problemAe} conflicts with the control bounds \eqref{problemAc} and \eqref{problemAd} \cite{xiao2021high,xiao2022sufficient}. To this end, the fixed $d_{\text{safe}}$ is modified to be a variable $d_{\text{min}}\leq d_{t}\leq d_{\text{max}}$ 
($d_{\text{min}}$ and $d_{\text{max}}$ are the lower and upper bounds for $d_{t}$, respectively), which is encouraged towards a larger value by adding a norm distance regularizer $P(\{d_{t}\})=-\eta\sum_{t=0}^H|d_{t}|$ ($\eta$ is a weighting factor) to the objective function
\eqref{problemAa}. This method automatically mitigates of conflicts between collision avoidance and control bound constraints.}

\section{Experiments}\label{section5} 

\subsection{Implementation}

We implemented the proposed EARN system using Python in Robot Operating System (ROS).
The high-fidelity CARLA simulation platform \cite{carla} is used for evaluations, which adopts unreal engine for high-performance rendering. 
Our EARN system is connected to CARLA via ROS bridge \cite{ros-bridge} and data sharing is {achieved} via ROS communications, where the nodes publish or subscribe ROS topics that carry the sensory, state, or action information.
We simulate two outdoor scenarios and one indoor scenario.
All simulations are implemented on a Ubuntu workstation with a 
$3.7$\,GHZ AMD Ryzen 9 5900X CPU and an NVIDIA $3090$\,Ti GPU.

We also implement EARN in a real-world multi-robot platform, where each robot has four wheels and can adopt Ackermann or differential steering.
The robot named LIMO has a 2D lidar, an RGBD camera, and an onboard NVIDIA Jetson Nano computing platform for executing the SLAM and local motion planning packages. 
The edge server is a manually controlled wheel-legged robot, Direct Drive Tech (DDT) Diablo, which has a 3D livox lidar and an onboard NVIDIA Orin NX computing platform for executing the SLAM and edge motion planning packages.
The Diablo is also equipped with a wireless access point. 

\begin{figure}[t]
    \centering
    \begin{subfigure}[t]{0.24\textwidth}
      \includegraphics[width=\textwidth]{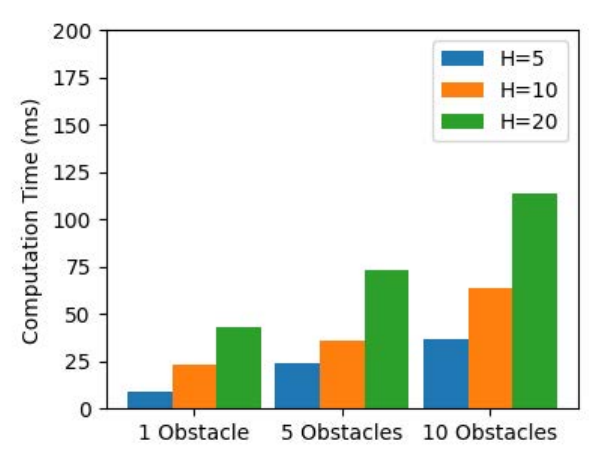}
        \caption{AMD Ryzen 9}
    \end{subfigure}
    \begin{subfigure}[t]{0.24\textwidth}
      \includegraphics[width=\textwidth]{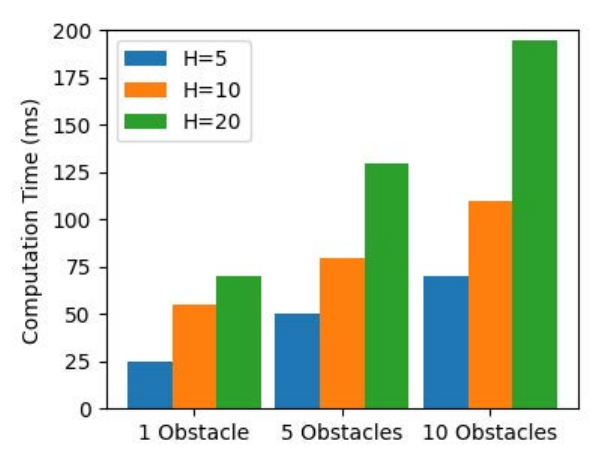}
        \caption{NVIDIA Orin NX}
    \end{subfigure}
         \vspace{-0.08in}
    \caption{Computation time (ms) of PDD planner versus the number of prediction horizons $H$ and obstacles $|\mathcal{M}(k)|$.    
    }
  \label{fig:comp_time}
        \vspace{-0.05in}
\end{figure}

\begin{figure}[t]
    \centering
    \begin{subfigure}[t]{0.23\textwidth}
      \includegraphics[width=\textwidth]{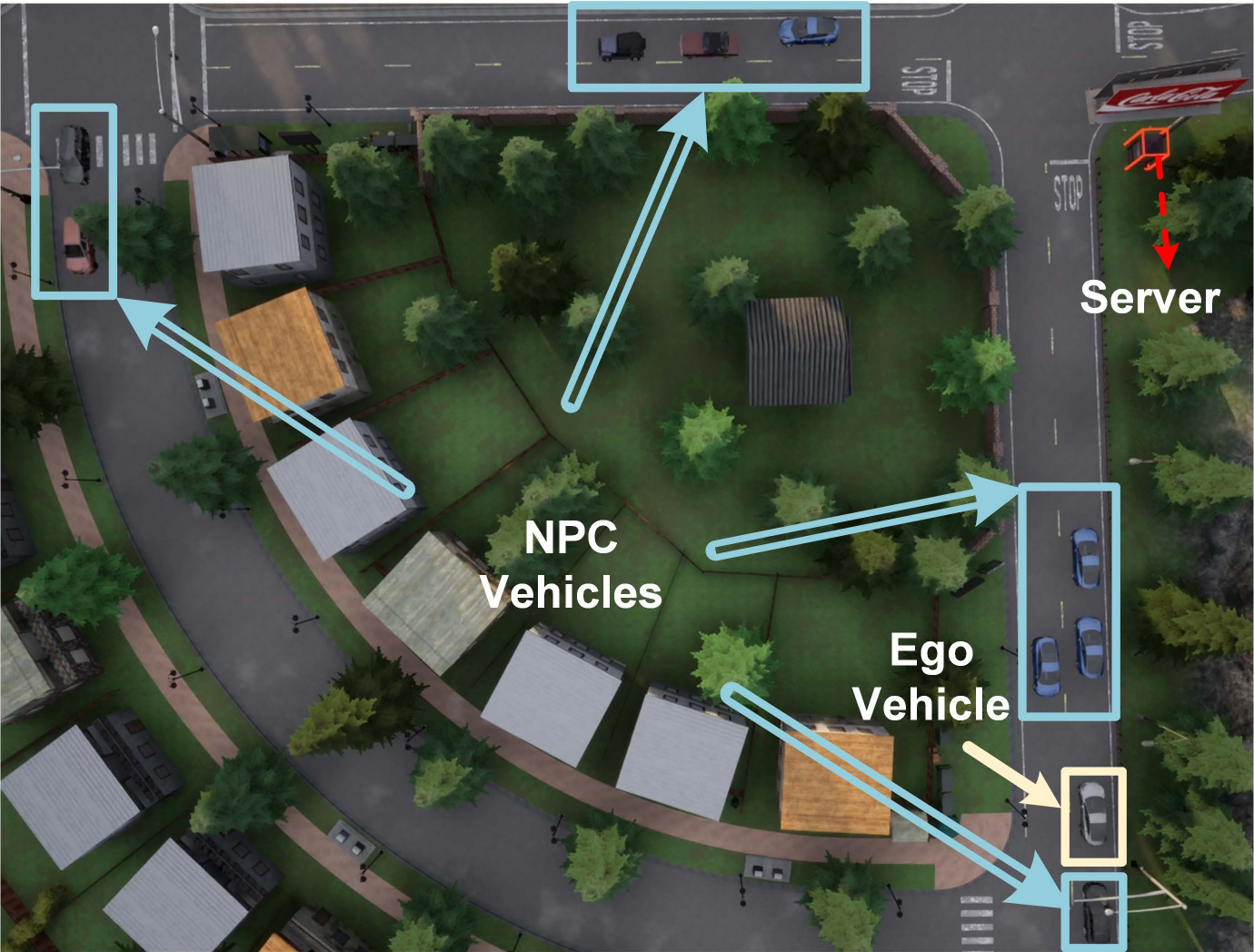}
      \caption{{Crossroad scenario in CARLA Town04 map}}
    \end{subfigure}%
    \quad 
    \begin{subfigure}[t]{0.23\textwidth}
      \includegraphics[width=\textwidth]{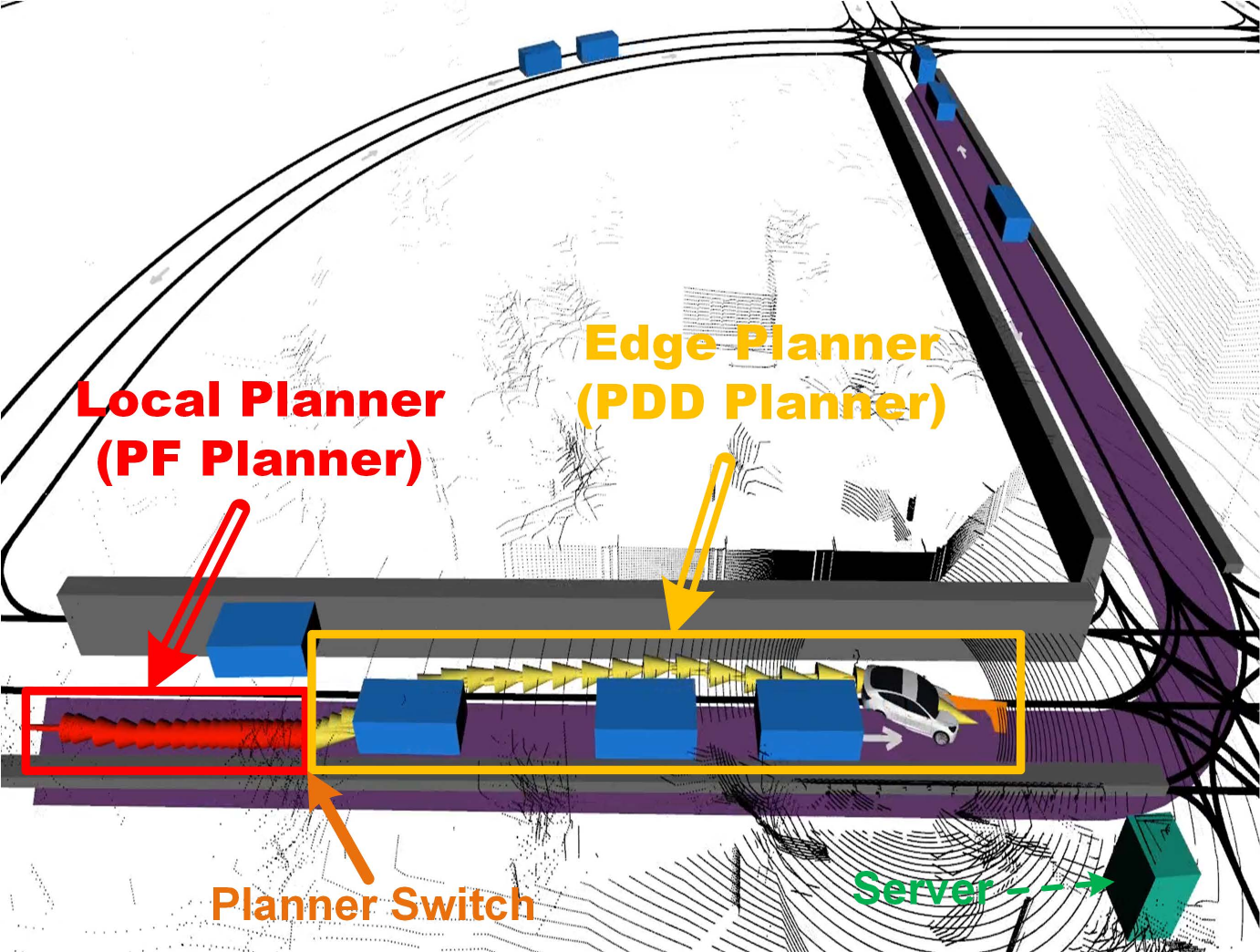}
        \caption{{EARN switches from PF to PDD for overtaking}}
    \end{subfigure}
     \vspace{-0.08in}
    \caption{{Crossroad scenario in CARLA Town04 and planner switching from PF to PDD performed by EARN.}
    }
    \label{fig:S1}
\end{figure}

\begin{figure*}[t]
    \centering
    \begin{subfigure}[t]{0.22\textwidth}
      \includegraphics[width=\textwidth]{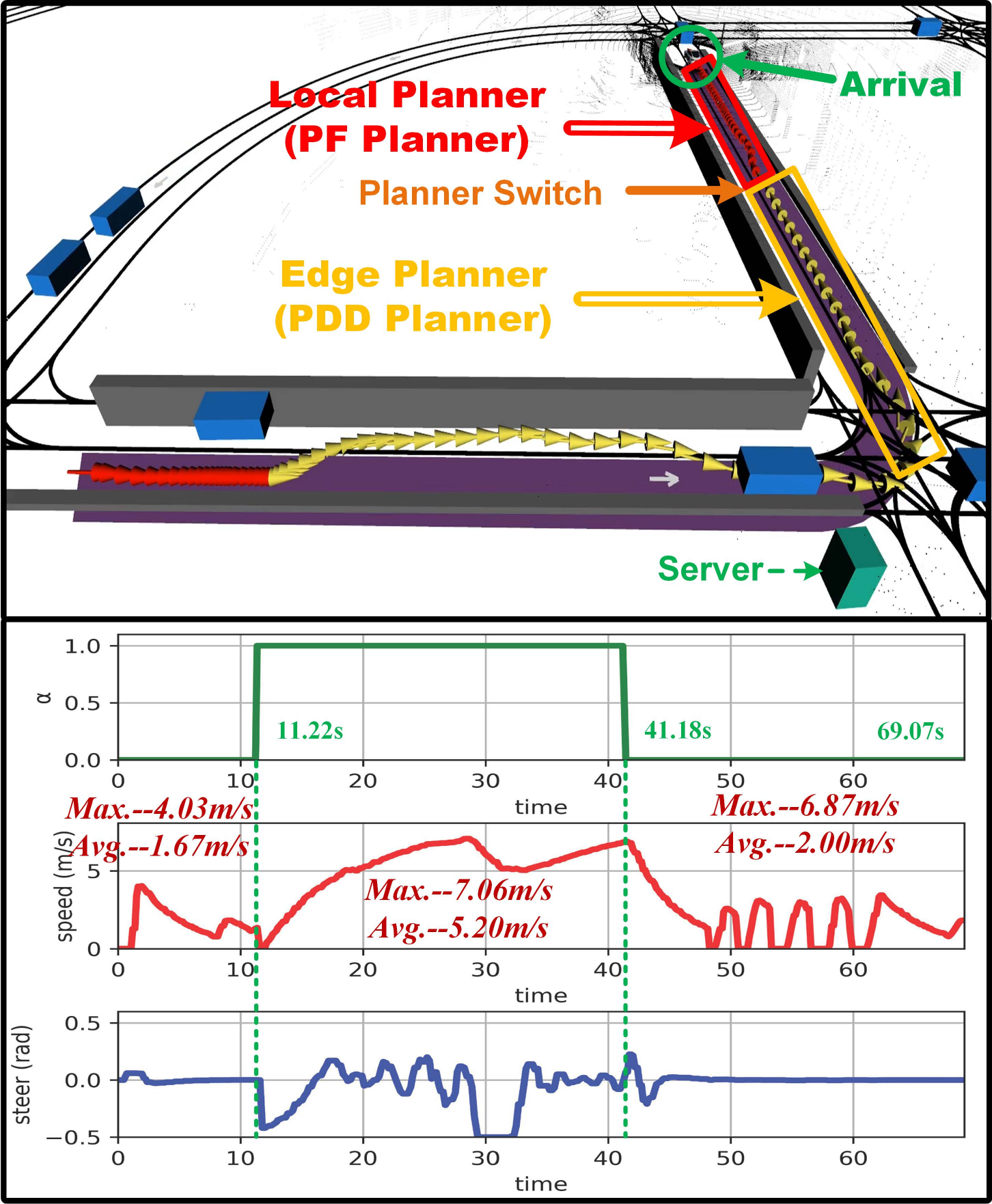}
        \caption{{EARN}}
    \end{subfigure}
    \quad
    \begin{subfigure}[t]{0.22\textwidth}
      \includegraphics[width=\textwidth]{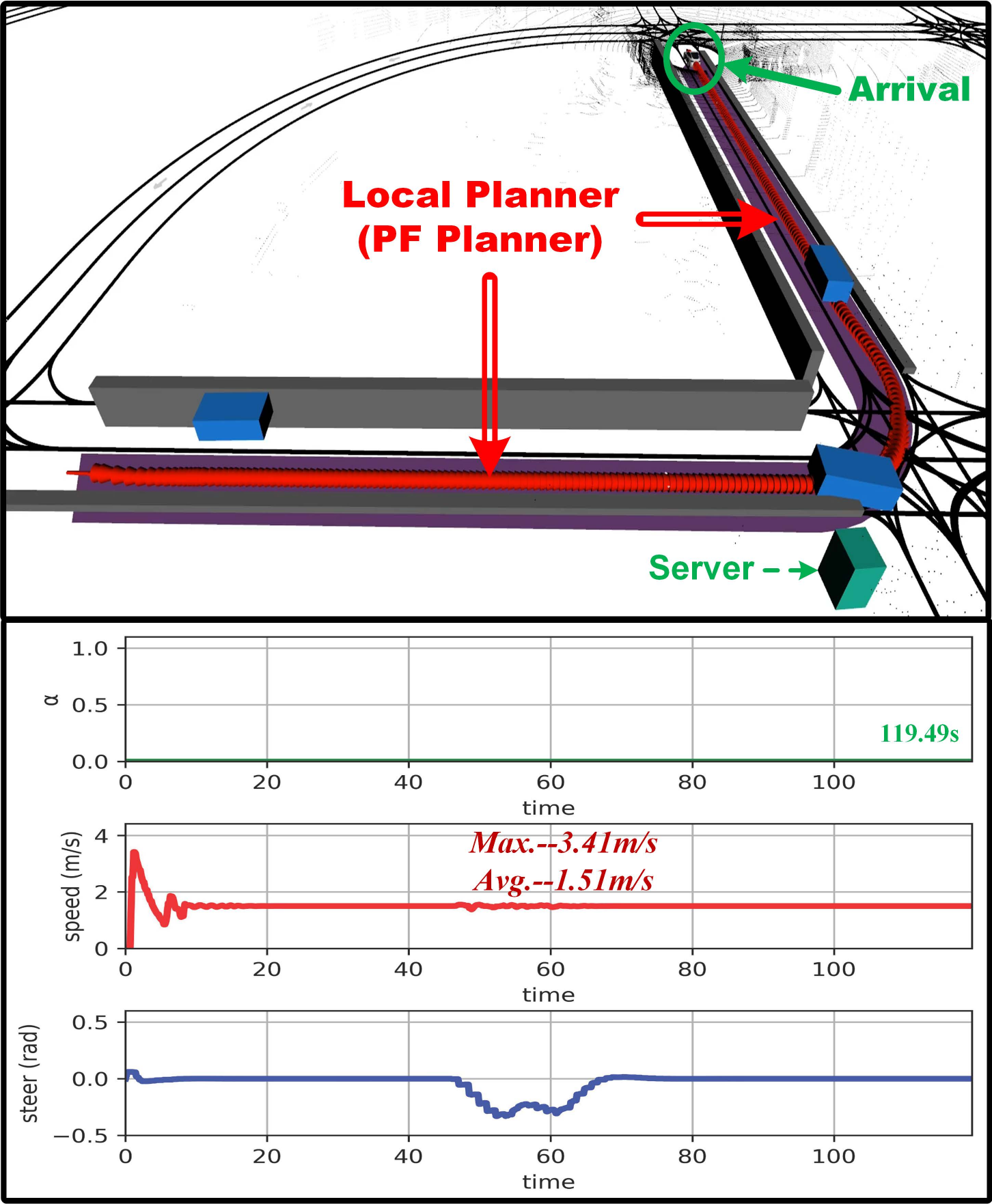}
      \caption{{PF}}
    \end{subfigure}%
    \quad 
    \begin{subfigure}[t]{0.22\textwidth}
      \includegraphics[width=\textwidth]{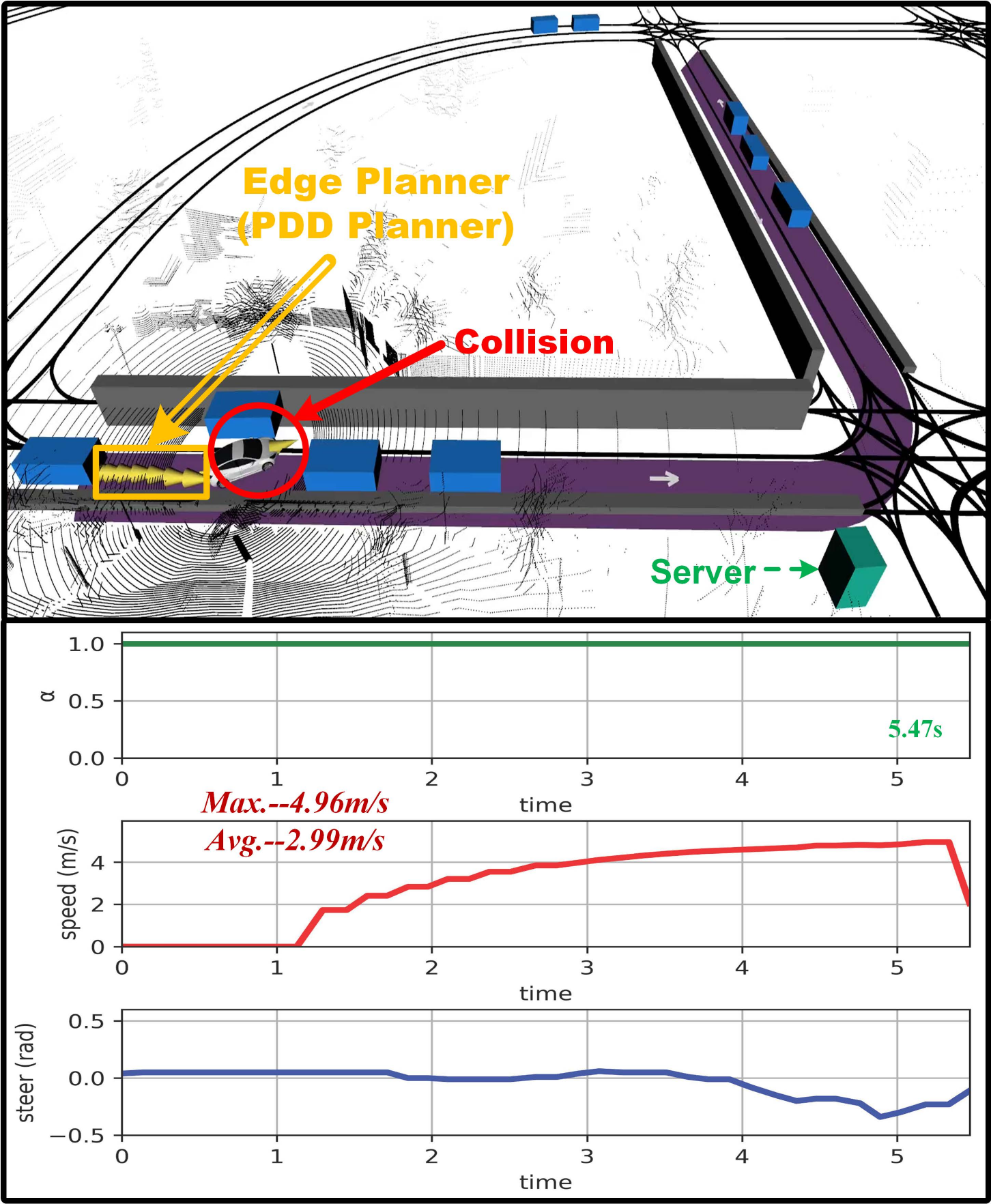}
        \caption{{PDD-Edge}}
    \end{subfigure}
    \quad 
    \begin{subfigure}[t]{0.22\textwidth}
      \includegraphics[width=\textwidth]{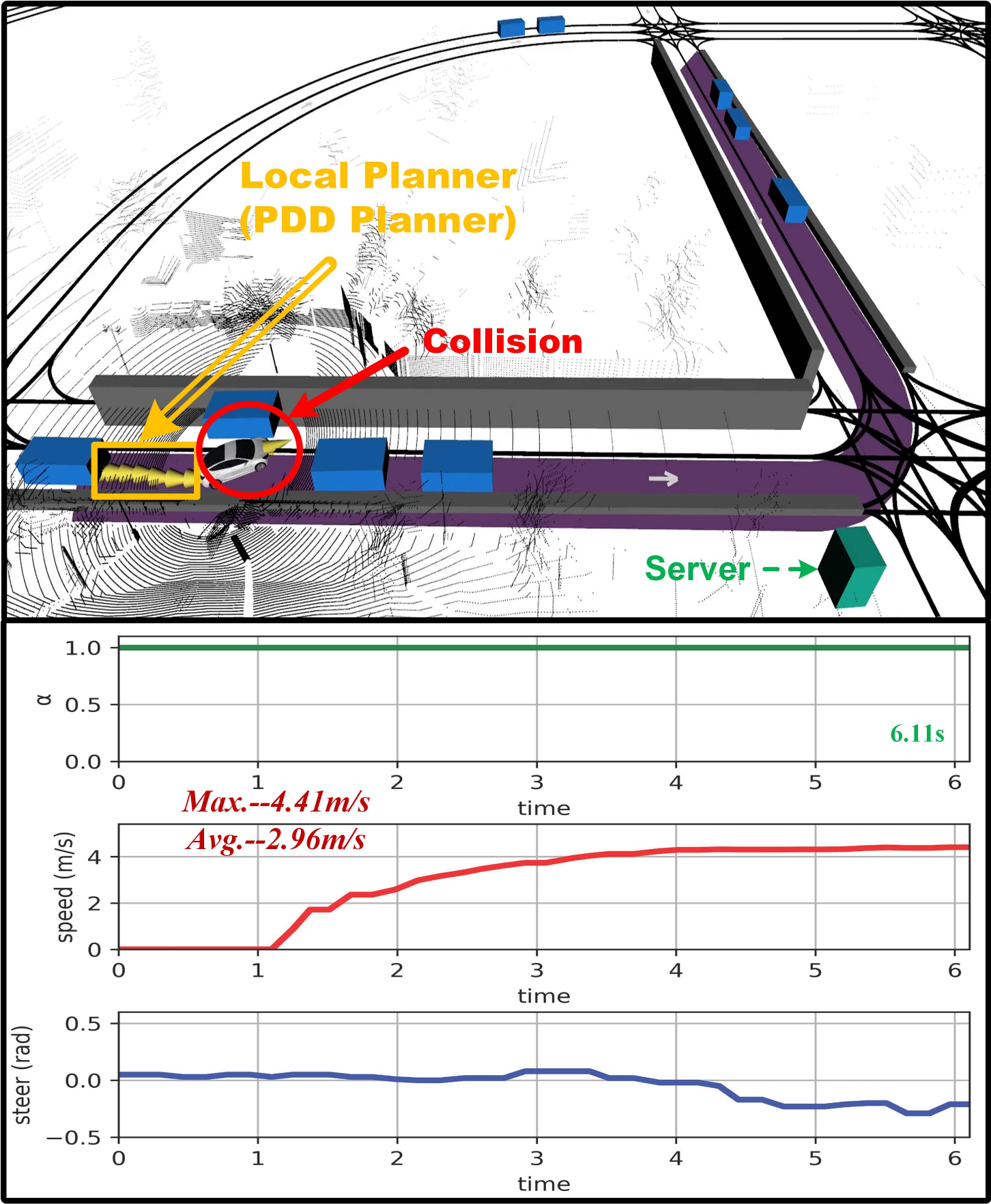}
        \caption{{PDD-Local}}
    \end{subfigure}
     \vspace{-0.08in}
    \caption{{Trajectories and control parameters of different schemes in crossroad scenario. Trajectories generated by the local and edge motion planners are marked in red and yellow, respectively.}
    }
    \label{fig:S1_trajectory}
          \vspace{-0.2in}
\end{figure*}

To obtain the parameters involved in model \eqref{comp_model}, we execute the PDD motion planning in Algorithm 2 on the 
AMD Ryzen 9 and NVIDIA Orin NX chips. The experimental data of computation time (ms) versus the number of prediction horizons $H$ and the number of obstacles $|\mathcal{M}(k)|$ is shown in Fig.~\ref{fig:comp_time}.
It can be seen that the computation time of PDD scales linearly with $H$ and $|\mathcal{M}_k|$, which corroborates the complexity analysis. As such, the value of parameter $p$ is set to 1. The computation time ranges from less than $10\,$ms to $200\,$ms, corresponding to a planning frequency of $5\,$Hz to over $100\,$Hz.
The parameters $\gamma,\tau$ are obtained by fitting the function $C_k=\gamma H|\mathcal{M}_k|+\tau$ to the experimental data in Fig.~\ref{fig:comp_time} using weighted least squares, and is given by $\gamma^*=0.6$ and $\tau^*=12\,$ms for 
ADM Ryzen 9 and 
$\gamma^*=1$ and $\tau^*=20\,$ms for NVIDIA Orin NX.

\subsection{Benchmarks}

We compare our method to the following baselines:
\begin{itemize}
\item[1)] Path Following (PF) planner \cite{wang2022extremum};
\item[2)] Collision avoidance MPC (CAMPC) \cite{jasontits}, which models each obstacle as a point and determines the collision condition by computing ${\bf{dist}}_c$;
\item[3)] RDA planner \cite{han2022rda}, which solves (19) in parallel via ADMM;
\item[4)] PDD-Edge, or PDD-E for short, which executes PDD planner at the edge server following the idea of \cite{edgeMPC}.
\item[5)] PDD-Local, or PDD-L for short, which executes PDD planner at the local robot.
    \item[6)] EDF, which is a collaborative motion planning scheme where robots are opportunistically selected for edge computing using a non-preemptive earliest deadline first (EDF) policy \cite{Syeda}.\footnote{Note that \cite{Syeda} does not consider motion planning and we combine our PDD planner with EDF for fair comparison.}
\end{itemize}

\begin{figure}[t]
    \centering
    \begin{subfigure}[t]{0.23\textwidth}
      \includegraphics[width=\textwidth]{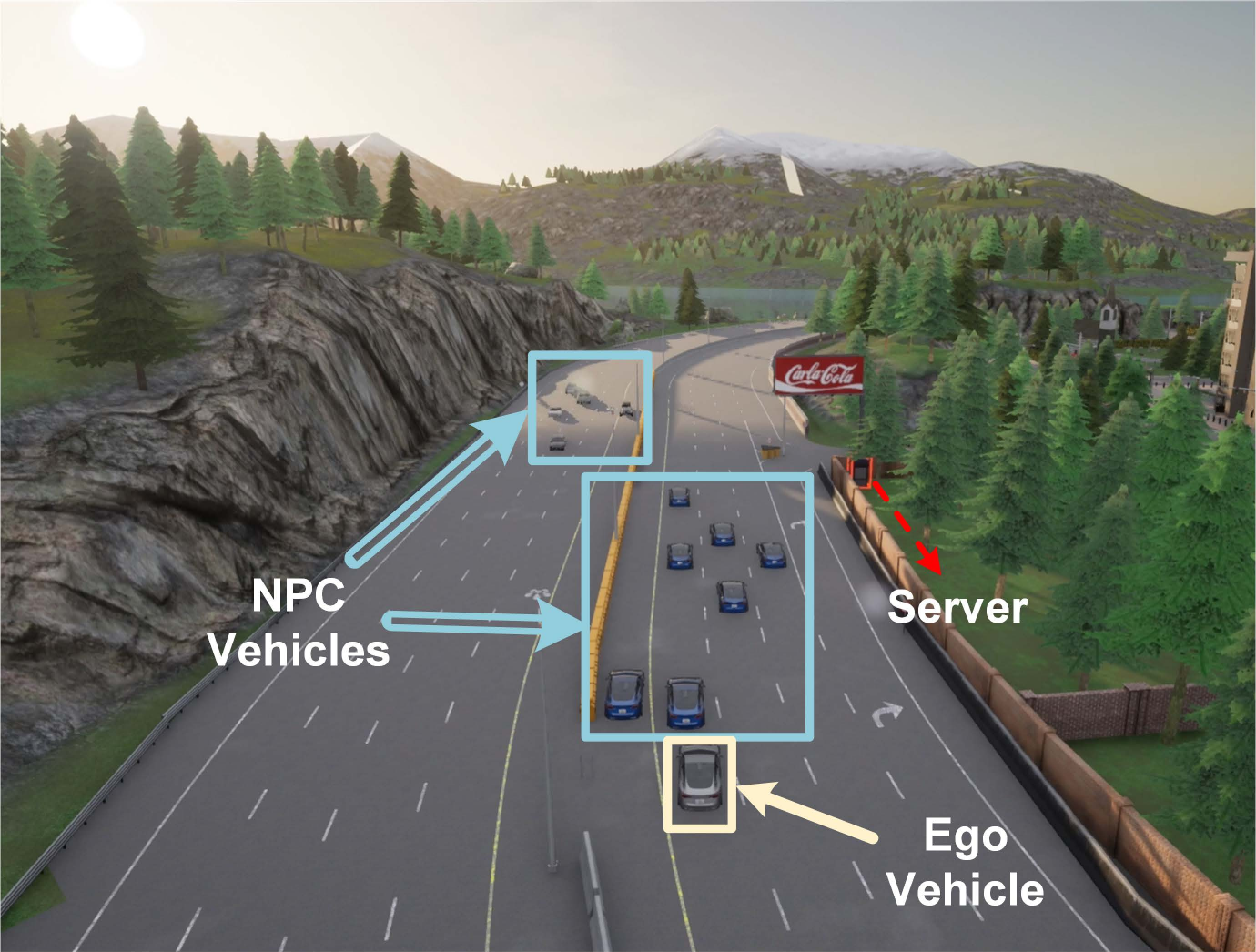}
      \caption{{Dense traffic scenario}}
    \end{subfigure}%
    \quad 
    \begin{subfigure}[t]{0.23\textwidth}
      \includegraphics[width=\textwidth]{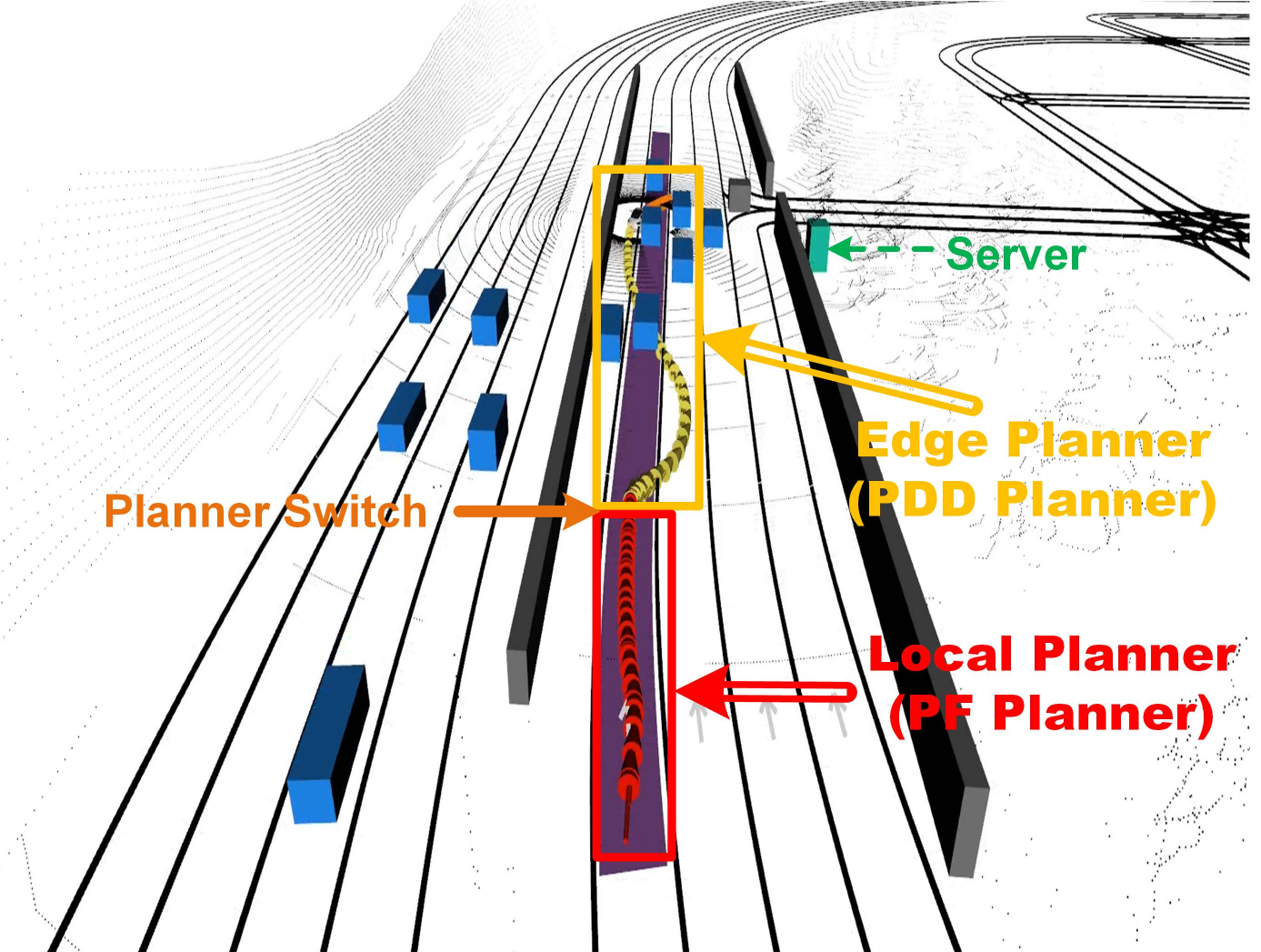}
        \caption{{Overtaking by PDD planner}}
    \end{subfigure}
     \vspace{-0.05in}
    \caption{{Dense traffic scenario in CARLA Town04 and overtaking performed by PDD planner.}
    }
    \label{fig:S2}
\end{figure}

\begin{figure*}[t]
    \centering
    \begin{subfigure}[t]{0.22\textwidth}
      \includegraphics[width=\textwidth]{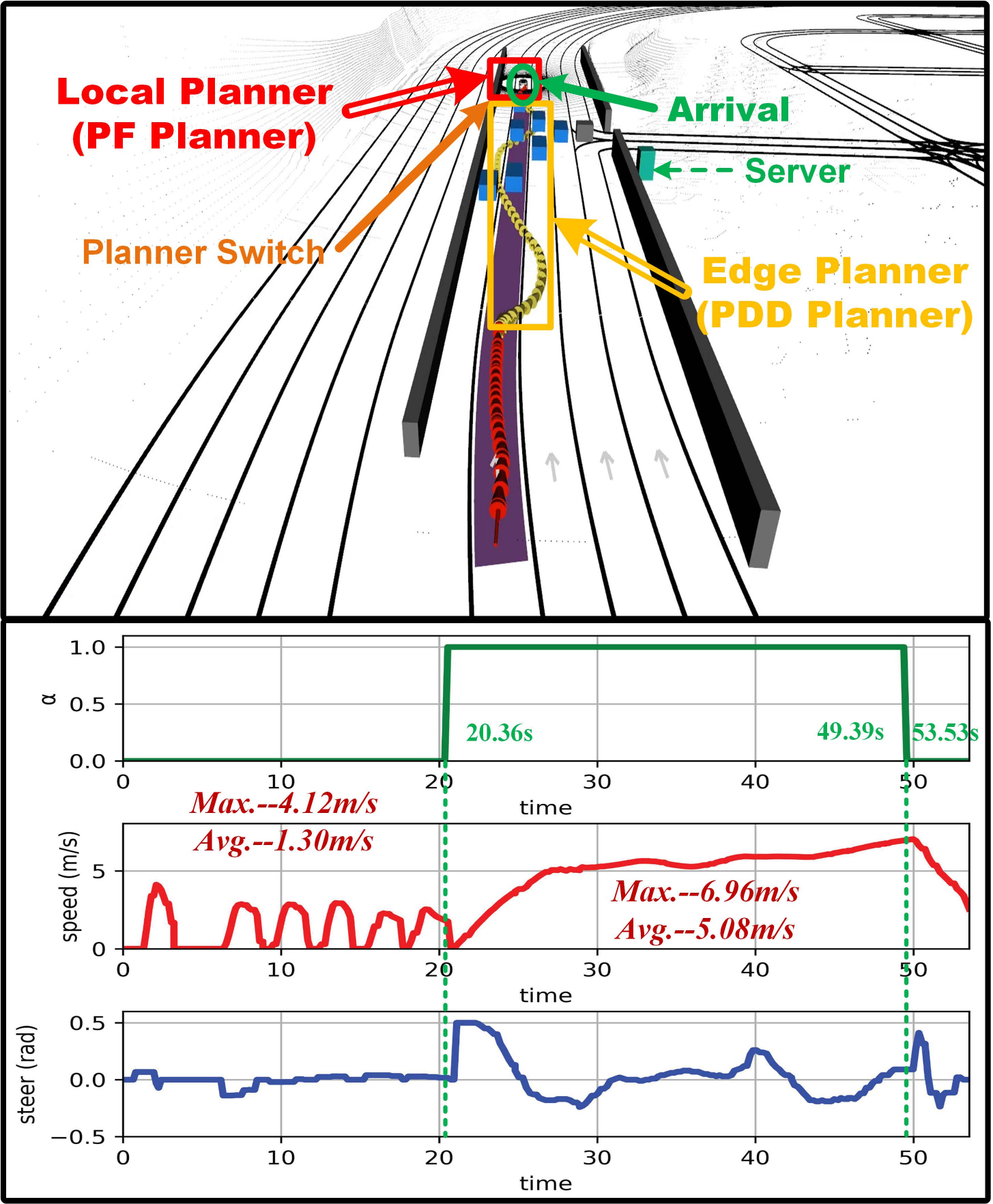}
        \caption{{PDD}}
    \end{subfigure}
    \quad
    \begin{subfigure}[t]{0.22\textwidth}
      \includegraphics[width=\textwidth]{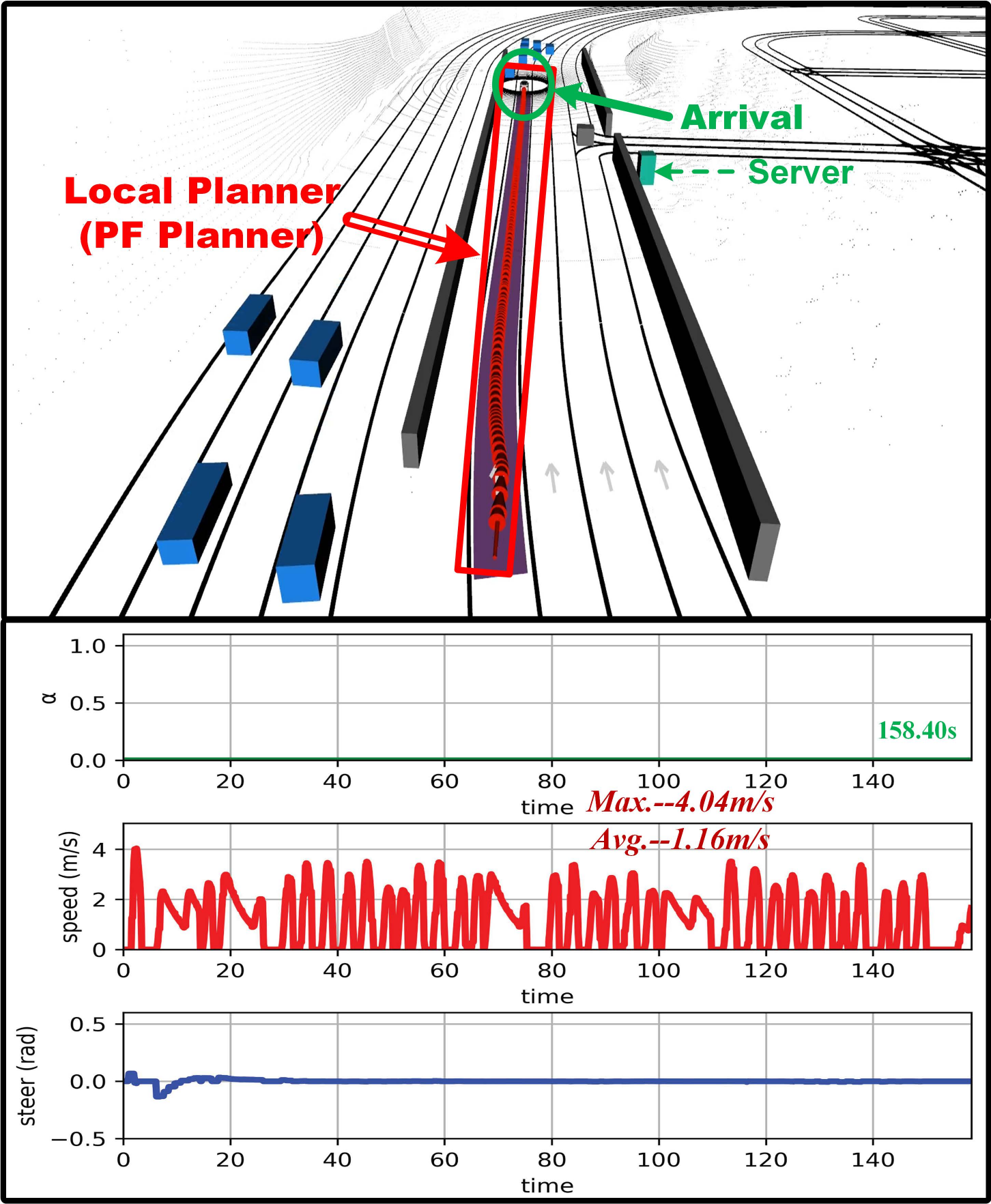}
      \caption{{PF}}
    \end{subfigure}
    \quad 
    \begin{subfigure}[t]{0.22\textwidth}
      \includegraphics[width=\textwidth]{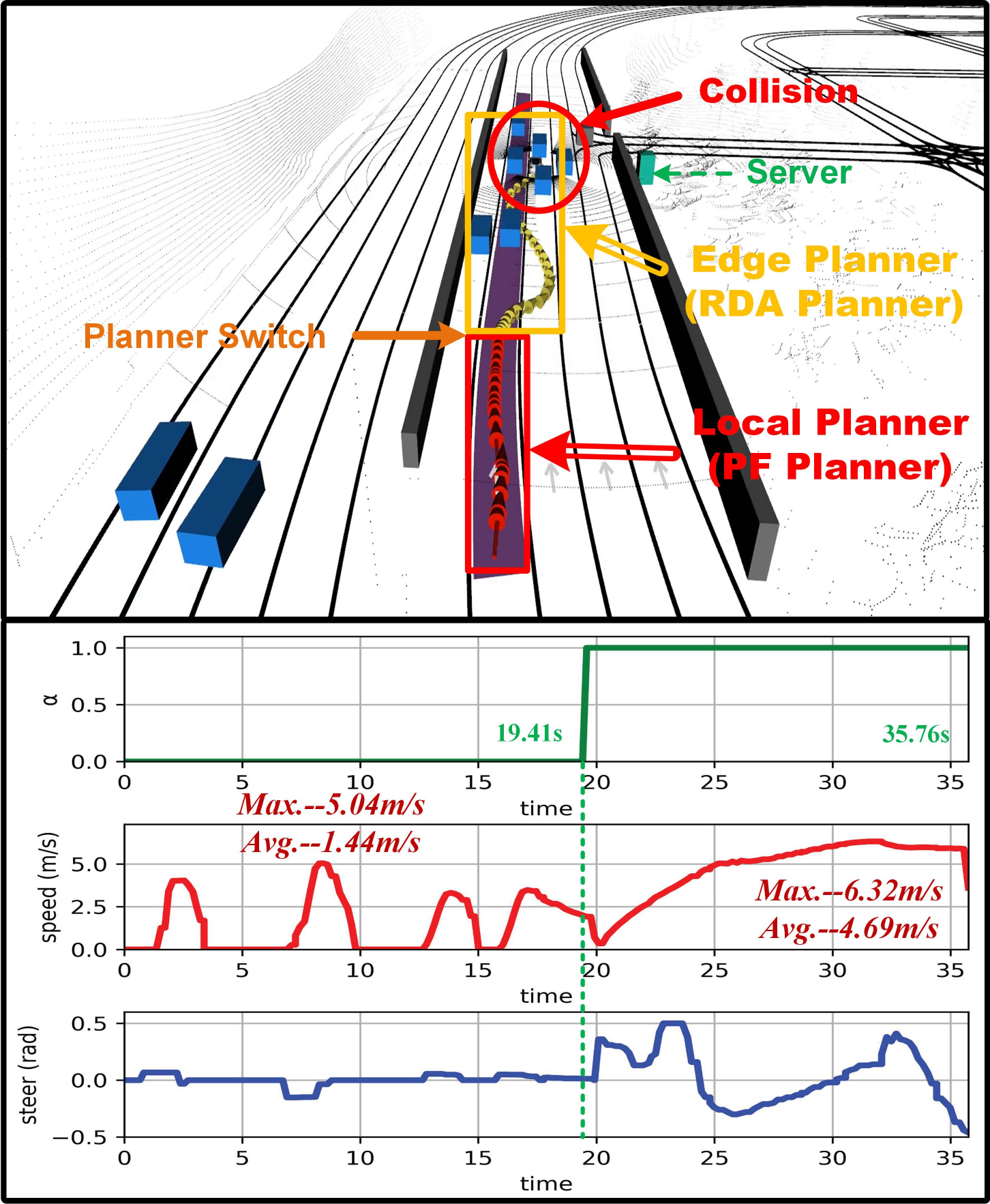}
        \caption{{RDA}}
    \end{subfigure}
    \quad 
    \begin{subfigure}[t]{0.22\textwidth}
      \includegraphics[width=\textwidth]{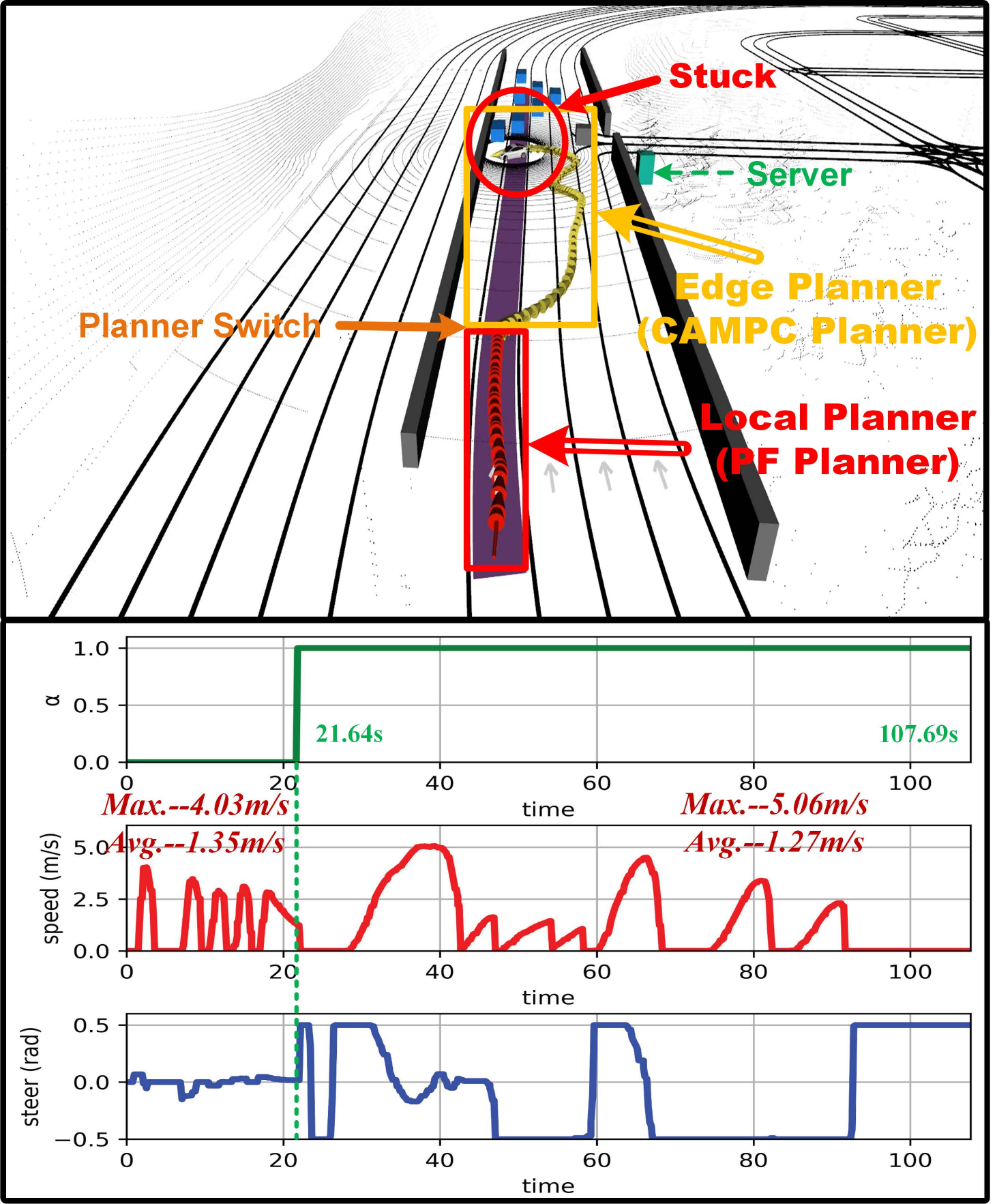}
        \caption{{CAMPC}}
    \end{subfigure}
          \vspace{-0.08in}
    \caption{{Trajectories and control parameters of different schemes in dense traffic scenario.}
    }
    \label{fig:S2_trajectory}
          \vspace{-0.15in}
\end{figure*}

\subsection{Single-Robot Simulation}

We first evaluate EARN in single-robot outdoor scenarios. 
The server is assumed to be equipped with AMD Ryzen 9.
The robot is assumed to be a low-cost logistic vehicle with its longitudinal wheelbase and lateral wheelbase being $2.87$\,m and $1.75$\,m, respectively. Based on these parameters, $(\mathbf{A}_{k,t}$, $\mathbf{B}_{k,t}$, $\mathbf{c}_{k,t})$ in \eqref{dynamics} can be calculated. 
The braking distance for planner $f_L$ is set to $d_{\mathrm{B}}=8.0$\,m and the safe distance for planner $f_E$ is set to $d_{\mathrm{safe}}=1$\,m.
The length of prediction horizon is set to $H=5$, with the time step between consecutive motion planning frames being $0.35$\,s.

We consider a crossroad scenario in CARLA Town04 map, where the associated location of edge server, the starting/goal position of robot (with a random deviation of $3\,$m), and the positions of non-player character (NPC) vehicles are shown in Fig. \ref{fig:S1}a. 
In this scenario, the maximum number of NPC vehicles in the local map $\mathcal{M}$ is $5$.
Using $H=5$ and $|\mathcal{M}|=5$, we have $C=27\,$ms when PDD is executed at the edge server. 
The computation time is assumed to be $200$ ms when executed locally at the ego vehicle.
The computation time of RDA is similar to that of PDD.
The computation time of PF is less than $20\,$ms.
The computation time of CAMPC is approximately $1/4$ of that of PDD, as CAMPC ignores the number of edges for each obstacle.
The computation latency threshold is set to $C_{\mathrm{th}}=50\,$ms in Algorithm 1 according to the operational speed in the outdoor scenario.
For this outdoor environment, we set $T_0\in\mathcal{U}(10,50)$ in ms if the distance from the vehicle to the server is smaller than $30\,$m, and $T_1\in\mathcal{U}(80,200)$ in ms otherwise \cite{mobisys}. 
The communication latency threshold is set to $D_{\mathrm{th}}=50\,$ms.

The trajectories and control parameters generated by EARN, PF, PDD-E and PDD-L are illustrated in Fig. \ref{fig:S1_trajectory}a--\ref{fig:S1_trajectory}d. It is observed that both EARN and PF planner can navigate the robot to the destination without any collision, while PDD-E and PDD-L lead to collisions at $t=5.47\,$s and $t=6.11\,$s, respectively. Moreover, EARN executes the planner switching at $t=11.22\,$s as observed in Fig.~\ref{fig:S1_trajectory}a. 
This empowers the robot with the capability to overtake front low-speed obstacles and increase the maximum speed from $4.03\,$m/s to $7.06\,$m/s and the average speed from $1.67\,$m/s to $5.20\,$m/s, by leveraging edge motion planning through low-latency communication access. 
In contrast, the maximum speed and average speed achieved by the PF planner are merely $3.41\,$m/s and $1.51\,$m/s, respectively.

\begin{table}[t]
    \centering
    \caption{Comparison of navigation time and success rate}
    \vspace{-0.03in}
        \scalebox{0.88}{
    \begin{tabular}{|c|c|c|c|c|}
        \hline
        \multirow{2}{*}{\diagbox{Metric}{Method}} & \multirow{2}{*}{EARN} & \multirow{2}{*}{PF} & \multirow{2}{*}{PDD-E} & \multirow{2}{*}{PDD-L}
        \\ &  &  & & \\
        \hline
        \multirow{2}{*}{Avg Navigation Time (s)} & \multirow{2}{*}{64.23} & \multirow{2}{*}{120.54} & \multirow{2}{*}{49.24} & \multirow{2}{*}{64.75} \\ &  &  &  &  \\\hline
        \multirow{2}{*}{Success Rate (\%)} & \multirow{2}{*}{98\%} & \multirow{2}{*}{100\%} & \multirow{2}{*}{64\%} & \multirow{2}{*}{58\%}
        \\
         &  &  &  &  \\
        \hline
    \end{tabular}
    \label{table_II}
        }
\end{table}

The quantitative comparisons are presented in Table \ref{table_II}, where the performance of each method is obtained by averaging $50$ trials. 
It can be seen that EARN reduces the average navigation time by $46.7\%$ compared to PF, as EARN can switch the planner adaptively and utilize the computing resources at the server as shown in Fig.~\ref{fig:S1}b. Moreover, both PDD-E and PDD-L suffer from low success rate.\footnote{A successful navigation requires the robot to reach the goal without any collision.} This is because PDD-E and PDD-L involve either computation or communication latency, resulting in low end-to-end planning frequency. The proposed EARN achieves a success rate significantly higher than those of PDD-E and PDD-L (98\% versus 64\% and 58\%), which demonstrates the necessity of planner switching performed by EARN. 

We also consider a dense traffic scenario in CARLA Town04 map as shown in Fig.~\ref{fig:S2}a with tens of dynamic obstacles. This experiment is used to evaluate the robustness of EARN in dynamic and complex environments. The trajectories and control vectors generated by PDD, PF, RDA, and CAMPC are illustrated in Fig.~\ref{fig:S2_trajectory}a--Fig.~\ref{fig:S2_trajectory}d. In this scenario, the PDD planner successfully navigates the robot to the destination in merely $53.53\,$s with average speed $5.08\,$m/s as shown in Fig.~\ref{fig:S2_trajectory}a. In contrast, the PF planner, while also finishing the task, costs over $158.40\,$s (with speed from $0\,$m/s to $4.04\,$m/s) due to its conservative navigation strategy. Moreover, the robot with RDA planner collides with NPC vehicle as the collision avoidance constraint is not strictly satisfied due to the fixed value of $\rho$. The CAMPC planner is neither able to navigate the robot to the destination, i.e., the robot gets stuck behind the dense traffic flow due to the center point distance model. Since RDA and CAMPC planners cannot accomplish the navigation task, we only compare the navigation time of PDD and PF. It is found that the PDD planner reduces the navigation time by $66.2\%$ compared to the PF planner, while achieving the same success rate.
\vspace{-0.06in}
\subsection{Multi-Robot Simulation}

To verify the effectiveness of Algorithm 1, we implement EARN in a multi-robot indoor scenario shown in Fig.~\ref{fig:indoor_trajectory}a, 
where robot 1 navigates in an office room with its target path marked in blue, robot 2 navigates in a corridor with its target path marked in green, robot 3 navigates in a conference room with its target path marked in orange, and robot 4 navigates from a corridor to a lounge with its target path marked in red. The number of NPC robots, which are marked as red boxes, in local maps 1--4 are are 
$(|\mathcal{M}_1|,|\mathcal{M}_2|, |\mathcal{M}_3|,|\mathcal{M}_4|)=(3, 6, 3, 0)$, respectively. The server (marked as a pink box) is assumed to be a Diablo wheel-legged robot with NVIDIA Orin NX.
Each robot is simulated as an Ackerman steering car-like robot with their length, width, longitudinal wheelbase, and lateral wheelbase being $32.2\,$cm, $22.0\,$cm, $20.0\,$cm, and $17.5\,$cm, respectively, and the parameters $(\mathbf{A}_{k,t}$, $\mathbf{B}_{k,t}$, $\mathbf{c}_{k,t})$ in \eqref{dynamics} are computed accordingly.
The braking distance for planner $f_L$ is set to $d_{\mathrm{B}}=1.3$\,m and the safe distance for planner $f_E$ is set to $d_{\mathrm{safe}}=0.1$\,m.
The length of prediction horizon is set to $H=20$, with the time step between consecutive motion planning frames being $0.25$\,s.
Using $H=20$ and the above $\{|\mathcal{M}_k|\}$, we have $(C_1,C_2,C_3,C_4)=(80, 140, 80, 20)\,$ms. 
The total computation latency threshold is set to $C_{\mathrm{th}}=240\,$ms according to the operational speed in indoor environments.
The execution time of PDD is assumed to be unacceptable when executed locally at the robot.
For this indoor environment, we generate the radio map Fig.~\ref{fig:indoor_trajectory}b and divide Fig.~\ref{fig:indoor_trajectory}b into two regions (i.e., $J=2$): 1) the yellow and light green area $\mathcal{Z}_0$; and 2) the dark green area $\mathcal{Z}_1$. 
According to the measurements, 
the communication latency $T_0\sim\mathcal{U}(10,50)$ in ms for $\mathcal{Z}_0$, and 
$T_1\sim\mathcal{U}(80,120)$ in ms for $\mathcal{Z}_1$.

\begin{figure*}[t]
    \centering
    \begin{subfigure}[t]{0.68\textwidth}
      \includegraphics[width=\textwidth]{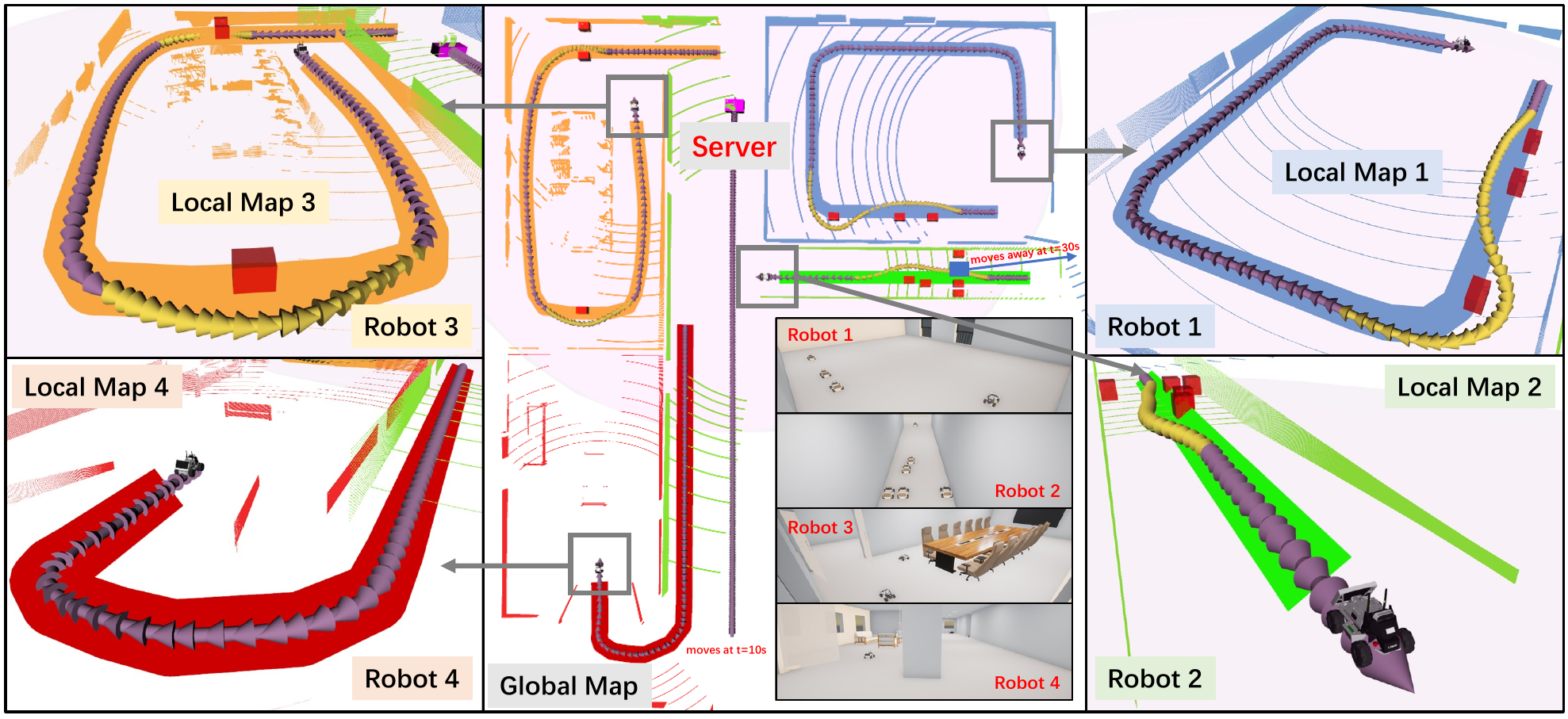}
      \caption{Multi-robot trajectories.}
    \end{subfigure}%
    ~
    \begin{subfigure}[t]{0.28\textwidth}
      \includegraphics[width=\textwidth]{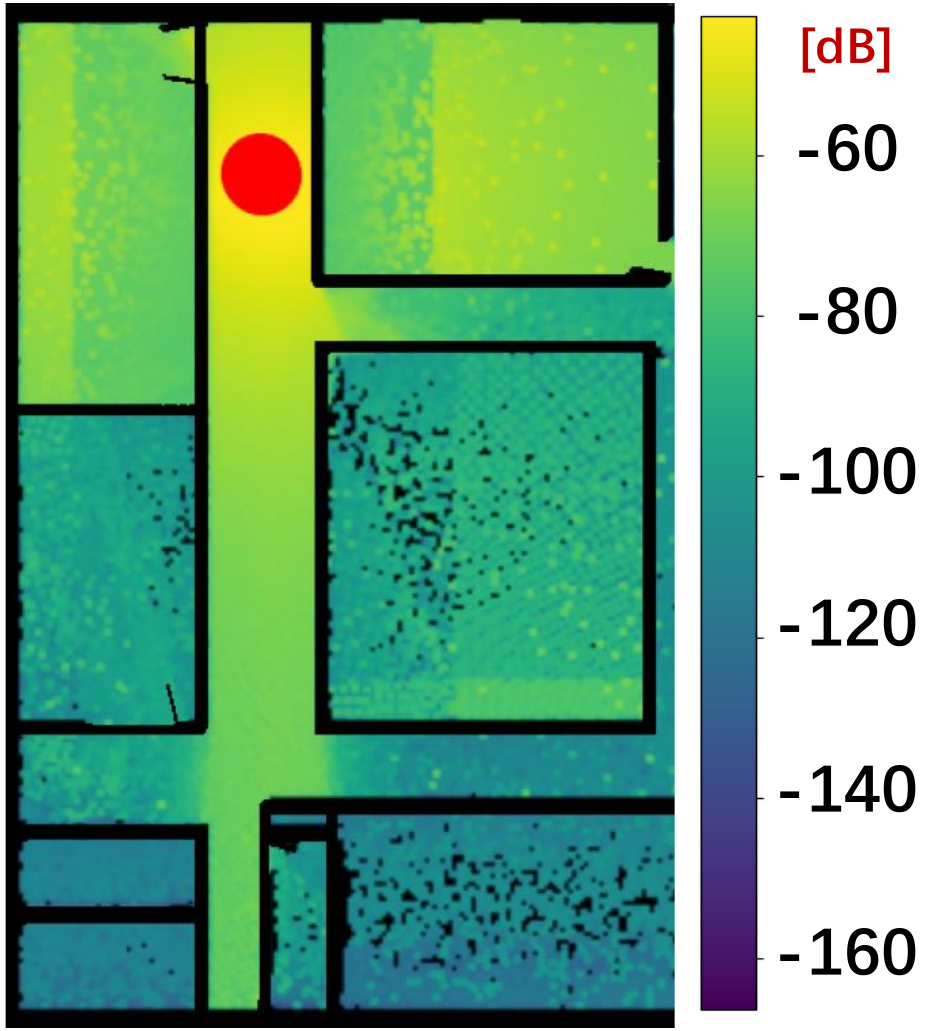}
        \caption{{Radio map for communication.}}
    \end{subfigure}
         \vspace{-0.08in}
    \caption{{Simulation results of the multi-robot indoor scenario: (a) Trajectories generated by the local (marked in purple) and edge (marked in yellow) motion planners; (b) Radio map of the server.}
    }
     \vspace{-0.1in}
    \label{fig:indoor_trajectory}
          \vspace{-0.05in}
\end{figure*}

\begin{figure*}[t]
    \centering
    \begin{subfigure}[t]{0.2\textwidth}
      \includegraphics[width=\textwidth]{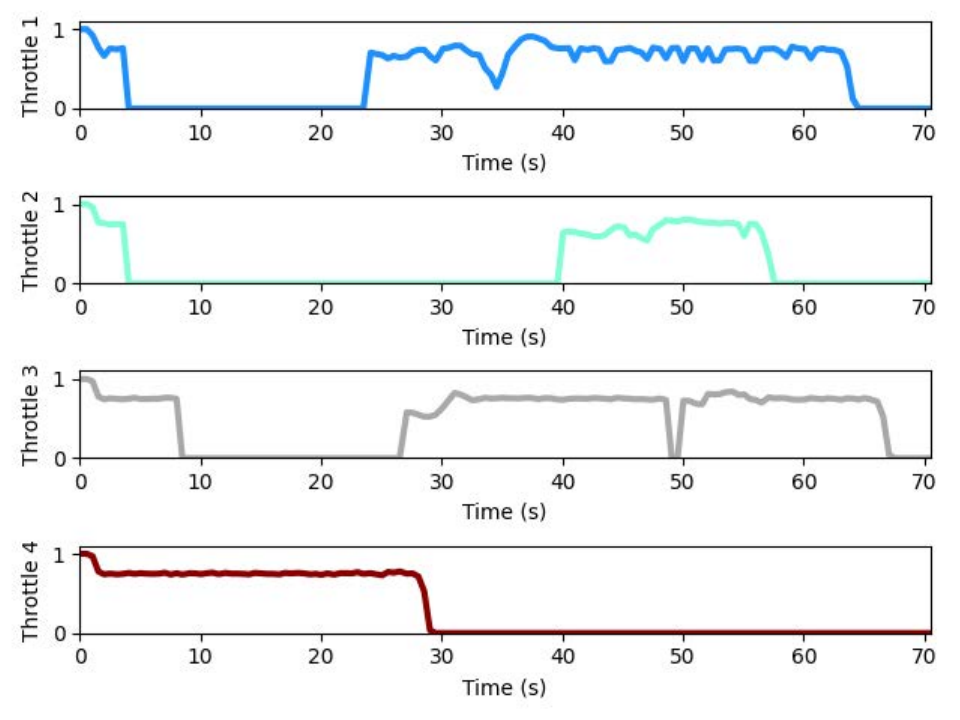}
      \caption{Throttle commands}
    \end{subfigure}%
    \quad 
    \begin{subfigure}[t]{0.2\textwidth}
      \includegraphics[width=\textwidth]{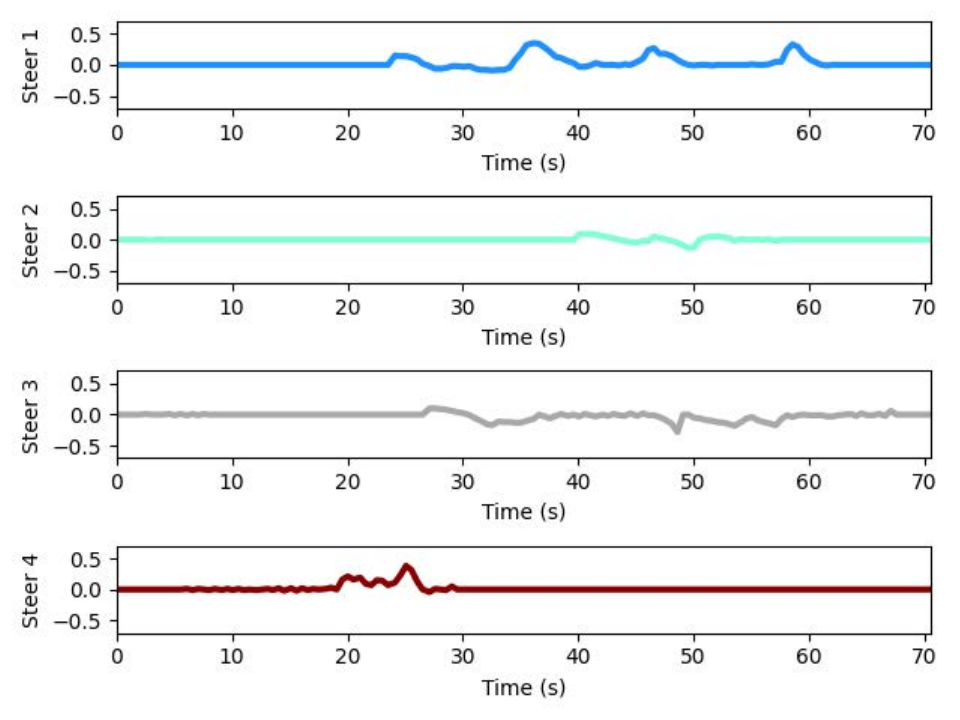}
        \caption{Steer commands}
    \end{subfigure}
    \quad 
    \begin{subfigure}[t]{0.2\textwidth}
      \includegraphics[width=\textwidth]{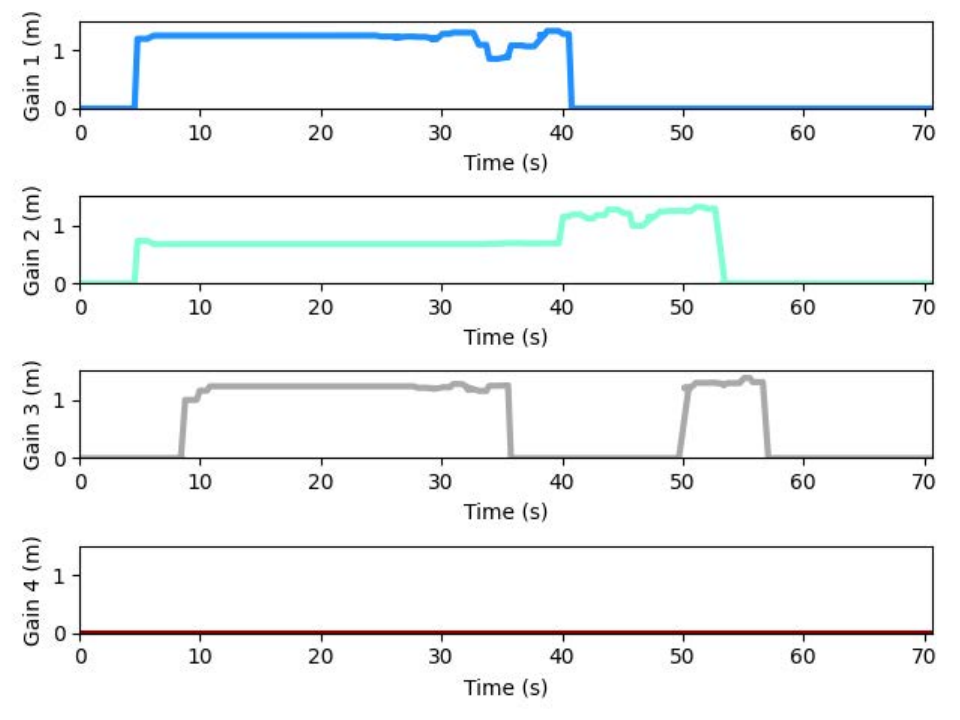}
        \caption{Switching gains}
    \end{subfigure}
    \quad 
    \begin{subfigure}[t]{0.2\textwidth}
      \includegraphics[width=\textwidth]{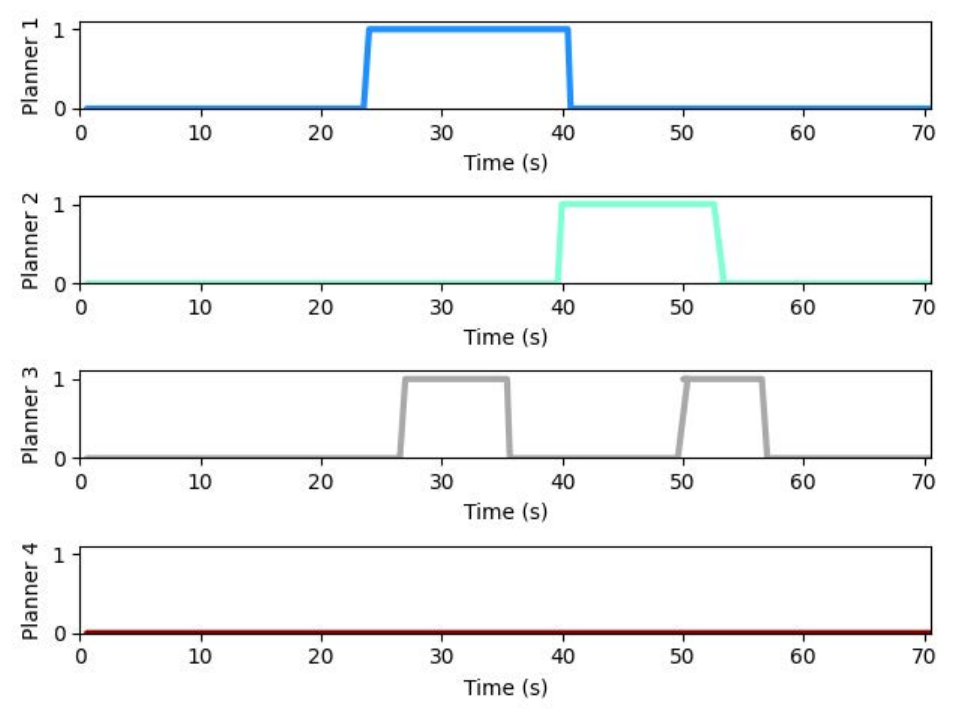}
        \caption{Planner selections}
    \end{subfigure}
    \vspace{-0.08in}
    \caption{The throttle commands, steer commands, switching gains, and planner selections of different robots.    
    }
    \label{fig:indoor_control}
          \vspace{-0.2in}
\end{figure*}

The trajectories of $4$ robots under the proposed EARN are shown in Fig.~\ref{fig:indoor_trajectory}a, where the purple trajectories are generated by local motion planner and the yellow trajectories are generated by edge motion planner.
The associated throttle commands, steer commands, switching gains $\{I_k\}$, and planner selections $\{\alpha_k\}$ are shown in Fig.~\ref{fig:indoor_control}a--\ref{fig:indoor_control}d.
In particular, starting from $t=0\,$s, all robots adopt local motion planning and move at a speed of $0.55\,$m/s.
Robot 4 encounters no obstacles along its target path; hence it keeps on moving using the local motion planner (as seen from the last subfigure of Fig.~\ref{fig:indoor_control}c--\ref{fig:indoor_control}d) until reaching the goal at $t=29.64\,$s (as seen from the last subfigure of Fig.~\ref{fig:indoor_control}a--\ref{fig:indoor_control}b).
In contrast, robots 1--3 stop in front of obstacles at $t=4.08\,$s, $3.88\,$s, and $8.52\,$s, respectively, as observed from Fig.~\ref{fig:indoor_control}a--\ref{fig:indoor_control}b.
Consequently, the switching gains of robots 1 and 3 increase from $0$ to $1.2$. However, they cannot switch their planners immediately, since the edge server is now at the other side of the map and the communication latency between the server and robot 1 or 3 would be large.
Note that the switching gain of robot 2 is only $0.67$, since the corridor is blocked by $4$ obstacles and the robot cannot pass the ``traffic jam'' even if it switches from local to edge planning.

\begin{table*}[t]
\centering
\caption{{Comparison of the robot navigation time under different computation time constraints}}
\label{comparison_gazebo}
\resizebox{\textwidth}{!}{
\begin{tabular}{cccccccccccc}
\cline{1-8}
\multicolumn{1}{c}{\multirow{2}{*}{\begin{tabular}[c]{@{}c@{}}Method \end{tabular}}} &
\multicolumn{1}{c}{\multirow{2}{*}{\begin{tabular}[c]{@{}c@{}}$C_{\mathrm{th}}$ (ms) \end{tabular}}} &
\multicolumn{1}{c}{\multirow{2}{*}{\begin{tabular}[c]{@{}c@{}} Num \end{tabular}}} &
\multicolumn{5}{c}{Navigation Time (s) $\downarrow$} & \\ \cline{4-8}  
\multicolumn{1}{c}{} & \multicolumn{1}{c}{} &  \multicolumn{1}{c}{} &
\multicolumn{1}{c}{Robot 1} &
\multicolumn{1}{c}{Robot 2} &
\multicolumn{1}{c}{Robot 3} &
\multicolumn{1}{l}{Robot 4} &
\multicolumn{1}{c}{Total} 
\\ \cline{1-8} 

EARN (ours)  & 240  & 4 robots &\textbf{64.80 (-1.5\%)}  &57.96  &\textbf{67.68 (-28.87\%)}  & 29.29 & \textbf{219.73 (-10.94\%)} \\
EDF \cite{Syeda}  & 240  & 4 robots &65.89  & \textbf{56.40 (-2.69\%)}  & 95.16  & 29.29 & 246.74 & \\ \cline{1-8} 
EARN (ours)  & 160 & 4 robots &\textbf{64.56 (-32.75\%)}  &66.72  &\textbf{80.04 (-40.81\%)}  & 29.29 &\textbf{240.61 (--23.96\%)} \\
EDF \cite{Syeda}  & 160 & 4 robots &96.00 &\textbf{55.92 (-16.18\%)} &135.24 & 29.29
&  316.45 \\ \cline{1-8} 
\end{tabular}%
\label{table_III}
}
\vspace{-0.2in}
\end{table*}



At $t=10\,$s, the server starts to move upwards at a speed of $0.88\,$m/s, and collaborates with robots 1 and 3 for collision avoidance using edge motion planning with a speed of $0.67\,$m/s at $t=24.12\,$s and $t=27.12\,$s, respectively. 
After completing the collision avoidance actions, the switching gains of robots 1 and 3 become $0$ and the planners are switched back to local ones to save computation resources at the server.
Furthermore, the switching gain of robot 2 increases from $0.67$ to $1.2$ at $t=40.08\,$s. 
This is because an NPC robot (marked as a blue box) in the corridor moves away at $t=30\,$s, leaving enough space for robot 2 to pass the traffic jam.

{The average navigation time of EARN and EDF is provided in Table \ref{table_III}.} 
The navigation time of robots 1--4 with EARN is $64.80\,$s, $57.96\,$s, $67.68\,$s, and $29.64\,$s, respectively. 
The navigation time of robots 1--4 with EDF is $65.89\,$s, $56.40\,$s, $95.16\,$s, and $29.29\,$s, respectively. 
Compared to EDF, the total navigation time of EARN is reduced by $12.1\%$.
This is because the EDF scheme is based on the expected task execution deadline, not on the low-level motion planning trajectories. As such, EDF fails to recognize the scenarios and traffic in real-time, leading to a potential resource-robot mismatch.
For instance, in the considered indoor scenario, robots 2 and 4 are expected to finish their navigation tasks in a shorter time, since their target paths are shorter than other robots as seen from Fig.~\ref{fig:indoor_trajectory}a. Therefore, EDF selects robots 2 and 4 for edge motion planning, despite the fact that robot 2 gets stuck in the traffic jam and robot 4 has a small switching gain from the local to edge planner.
The navigation time of EARN and EDF under $C_{\mathrm{th}}=160\,$ms is also provided in Table \ref{table_III}. Due to this tighter time constraint, the server cannot navigate multiple robots at the same time, and the waiting time for robots 1 and 3 becomes larger. 
Consequently, the robot navigation time increases.
However, the proposed EARN still outperforms EDF by a large margin.

\begin{figure}[!t]
    \centering
    \begin{subfigure}[t]{0.23\textwidth}
      \includegraphics[width=\textwidth]{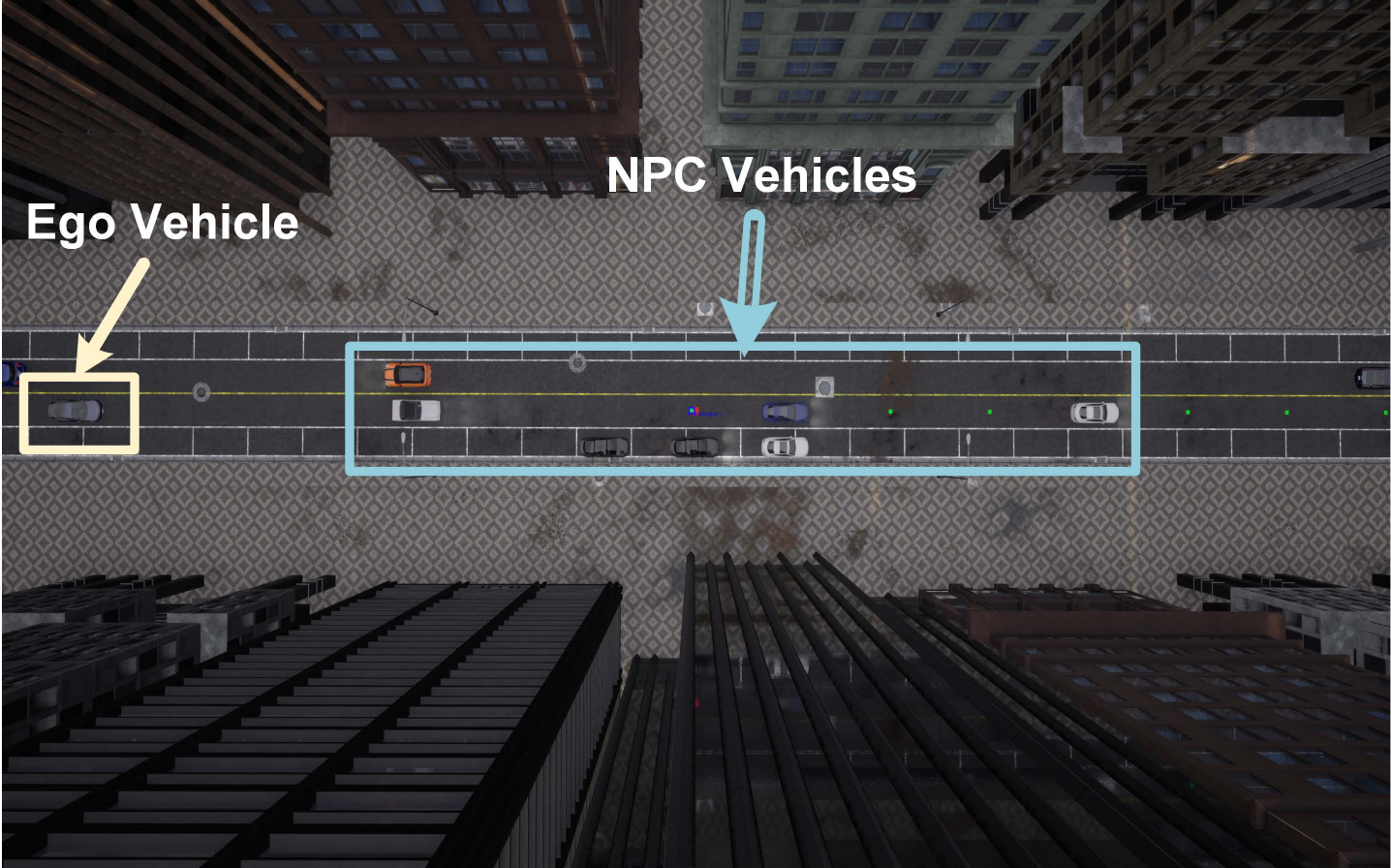}
      \caption{{Urban driving scenario}}
    \end{subfigure}%
    \quad 
    \begin{subfigure}[t]{0.23\textwidth}
      \includegraphics[width=\textwidth]{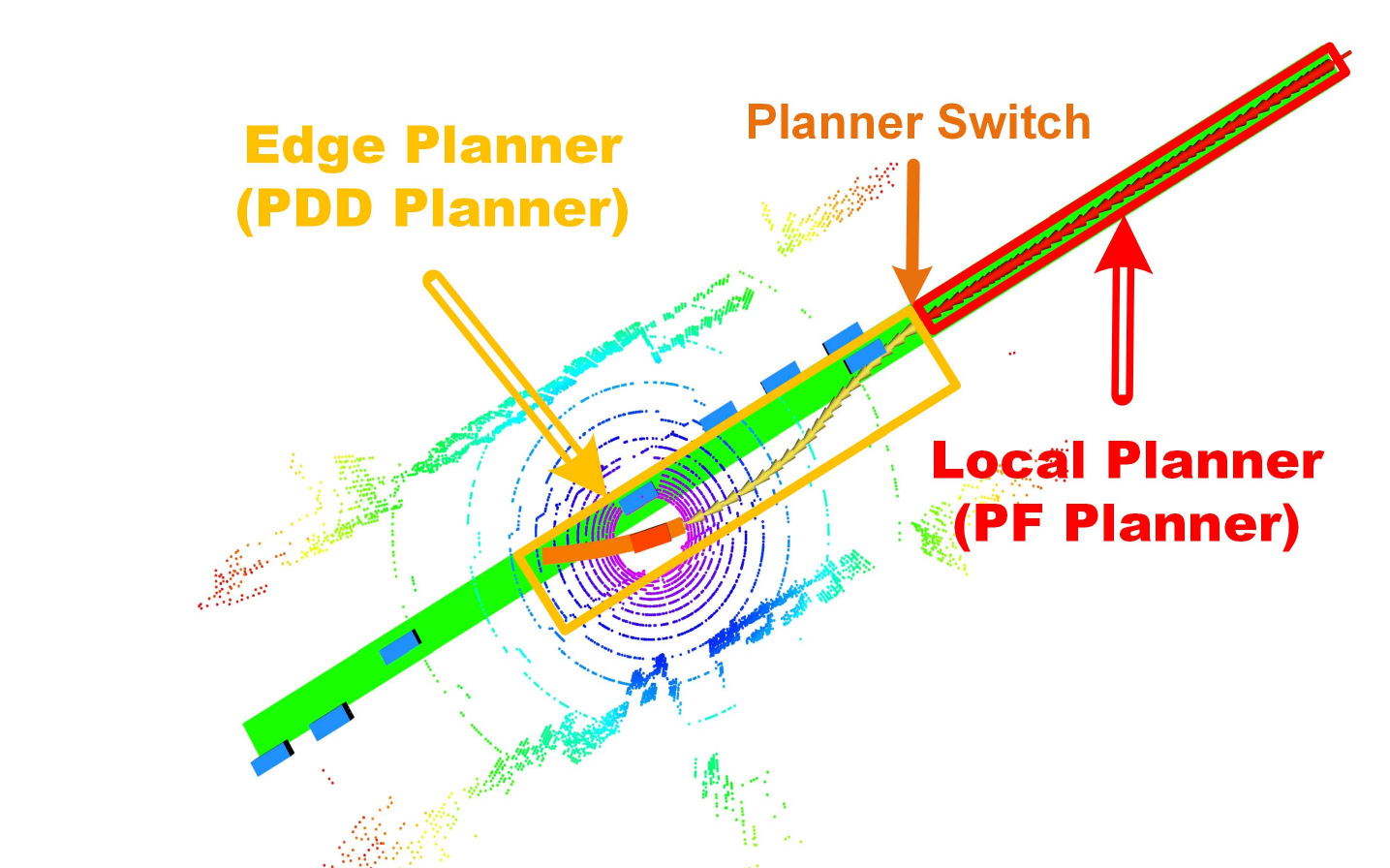}
        \caption{{Planner switched by EARN}}
    \end{subfigure}
    \caption{{Urban driving scenario in Intel CARLA Challenge.}
    }
    \label{fig:Challenge}
\end{figure}

\begin{figure}[t]
    \centering
    \begin{subfigure}[t]{0.22\textwidth}
      \includegraphics[width=\textwidth]{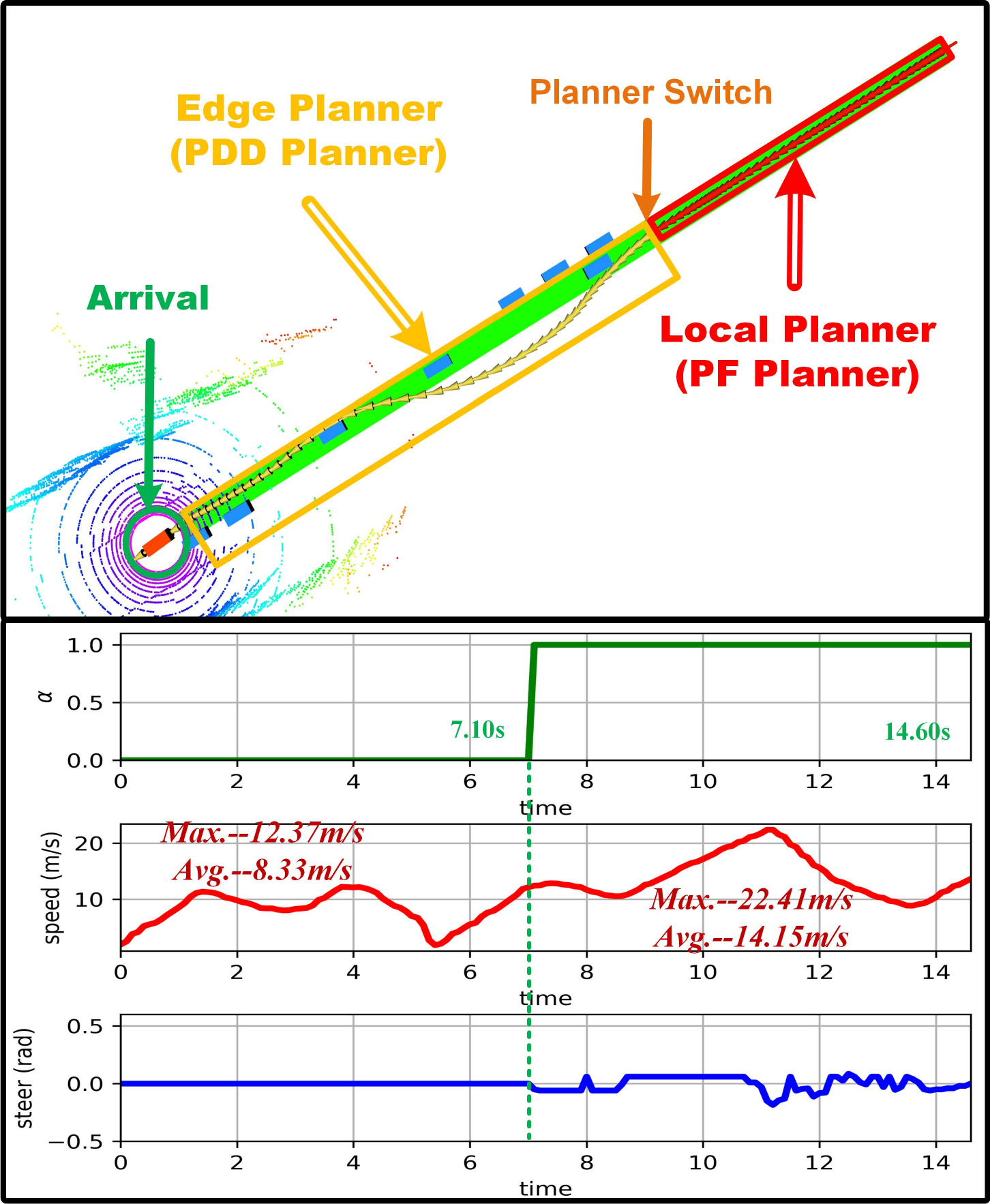}
      \caption{{EARN}}
    \end{subfigure}%
    \quad 
    \begin{subfigure}[t]{0.22\textwidth}
      \includegraphics[width=\textwidth]{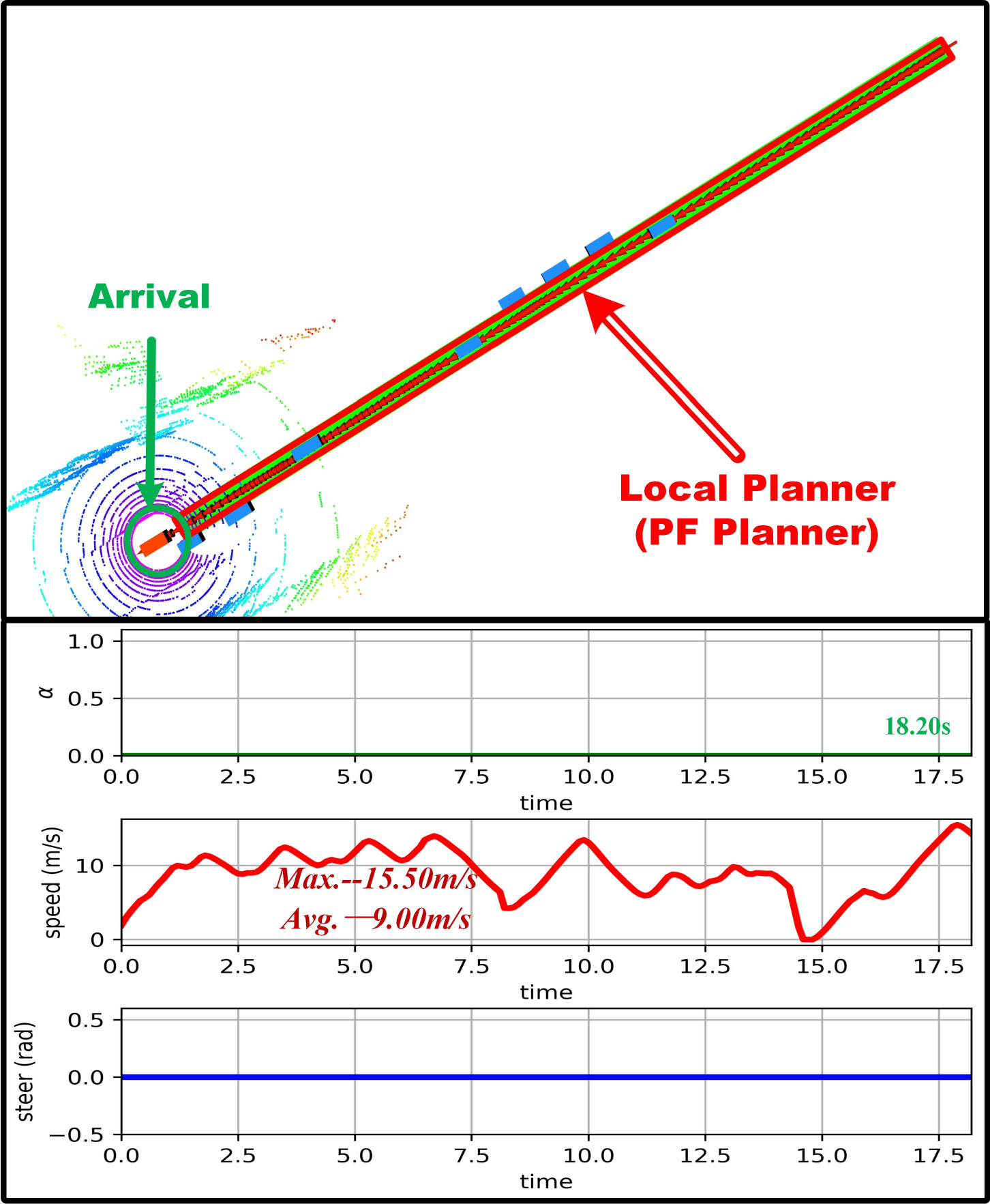}
        \caption{{PF}}
    \end{subfigure}
     \vspace{-0.08in}
    \caption{{Trajectories and control parameters of different schemes in the official Intel CARLA challenge.}}
    \label{fig:Challenge_trajectory}
    \vspace{-0.05in}
\end{figure}

\begin{figure}[t]
    \centering
      \includegraphics[width=0.46\textwidth]{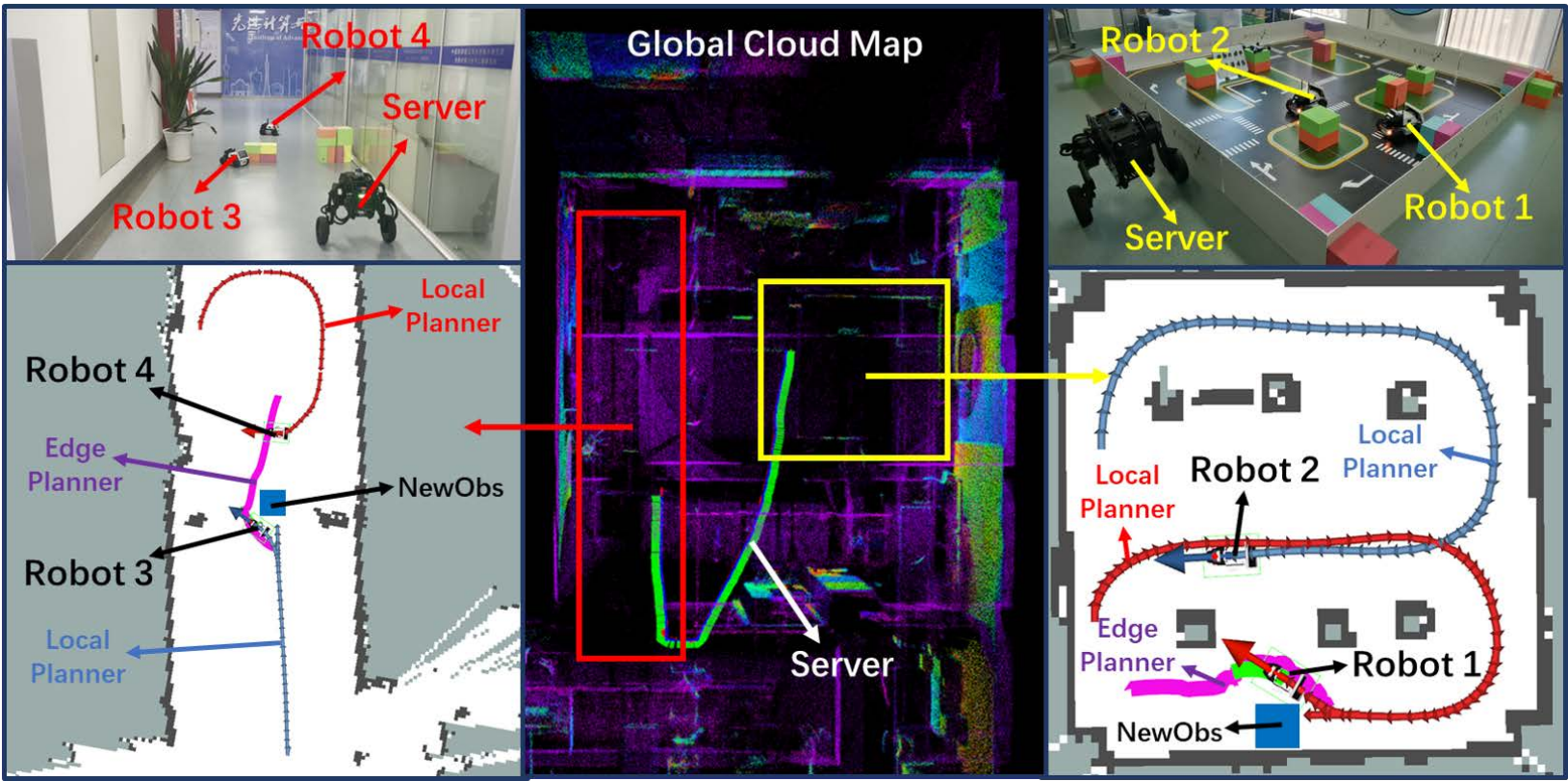}
    \caption{Real-world experiment 1.}
    \label{fig:real_exp}
    \vspace{-0.08in}
\end{figure}

{
\subsection{Intel CARLA Challenge}
The performance of EARN is also verified in the official Intel CARLA Challenge (\url{https://leaderboard.carla.org/challenge/ }). As illustrated in Fig. \ref{fig:Challenge}a, the competition is an urban driving scenario of the CARLA Town12 map, with various NPC vehicles on the road. It is observed from Fig. \ref{fig:Challenge}b that the robot accomplishes the overtaking action and reduces the navigation time by switching the adopted planner to PDD at $t=7.10$s. The trajectories and control parameters generated by EARN and PF are also illustrated in Fig. \ref{fig:Challenge_trajectory}a--\ref{fig:Challenge_trajectory}b. It is observed that both EARN and PF planners are able to navigate the robot to the destination without any collision. However, the navigation time of EARN (14.60s) is reduced by $19.78\%$ compared to that of the PF planner (18.20s), which reveals the speed boost and performance enhancement brought by EARN.}
\vspace{-0.1in}
\subsection{Real-World Testbed}

Finally, to verify the hardware-software compatibility of EARN and its robustness against sensor and actuator uncertainties, 
we implement EARN in a real-world testbed consisting of a Diablo wheel-legged robot server and $4$ LIMO car-like robots. 
As shown in Fig.~\ref{fig:real_exp}, we adopt the livox lidar and the fast-lio algorithm \cite{lio} to obtain a cm-level localization of the Diablo robot and the global cloud map of the indoor scenario. 
The trajectory of Diablo is marked as red-green-blue axes.
The cloud map consists of two regions, where robots 1 and 2 navigate inside the sandbox (marked by a yellow box) and robots 3 and 4 navigate in the corridor (marked by a red box).
All robots need to avoid collisions with the static environments as well as the cubes therein.

\begin{figure*}[t]
    \centering
    \begin{subfigure}[t]{0.23\textwidth}
      \includegraphics[width=\textwidth]{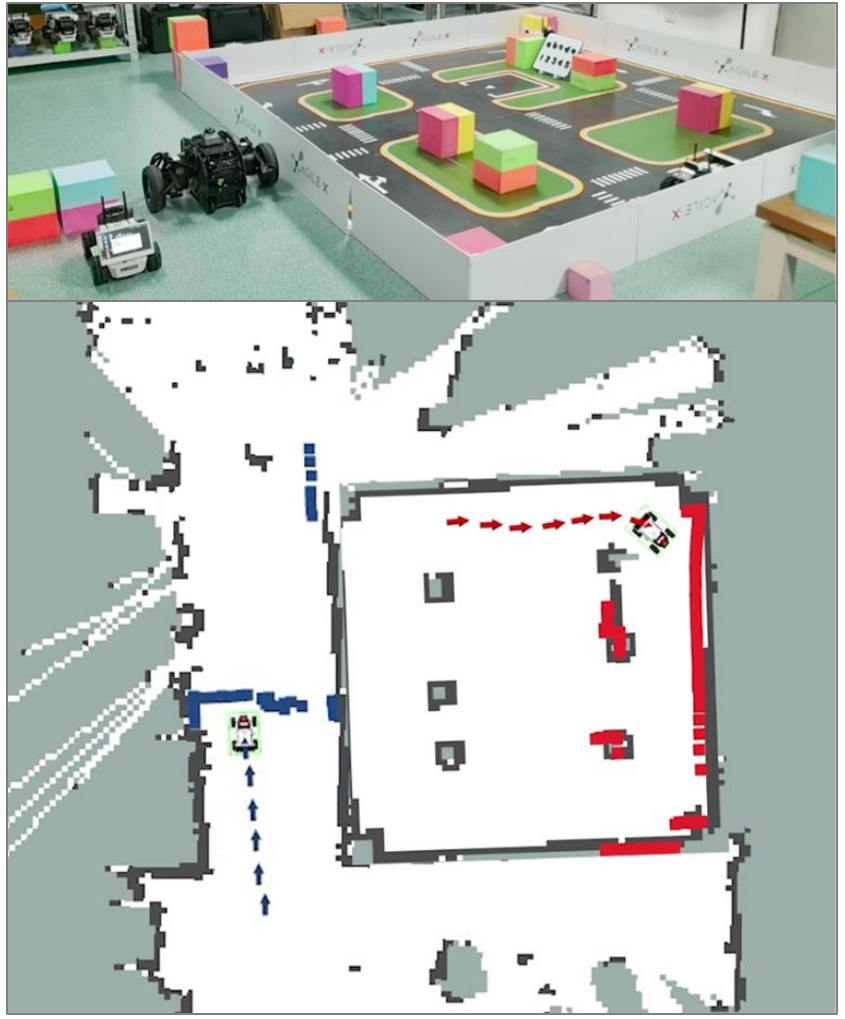}
      \caption{{No switching gain.}}
    \end{subfigure}%
    ~
    \begin{subfigure}[t]{0.23\textwidth}
      \includegraphics[width=\textwidth]{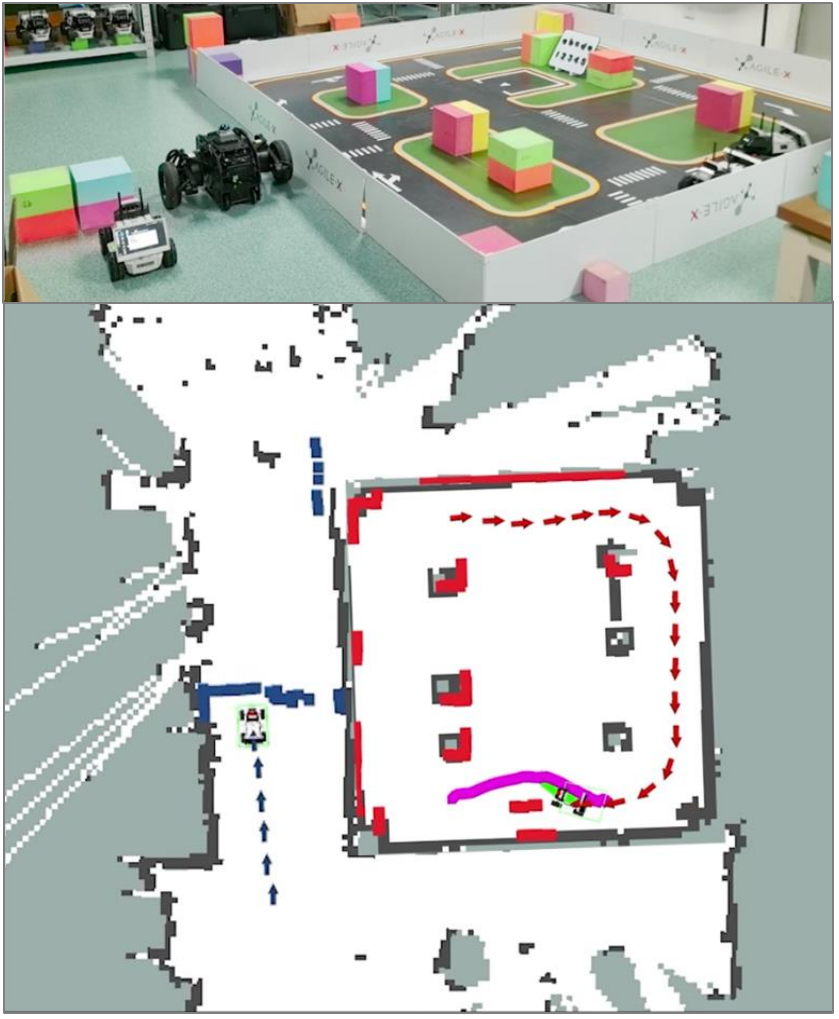}
        \caption{{Robot 1 switches.}}
    \end{subfigure}
    ~
        \begin{subfigure}[t]{0.23\textwidth}
      \includegraphics[width=\textwidth]{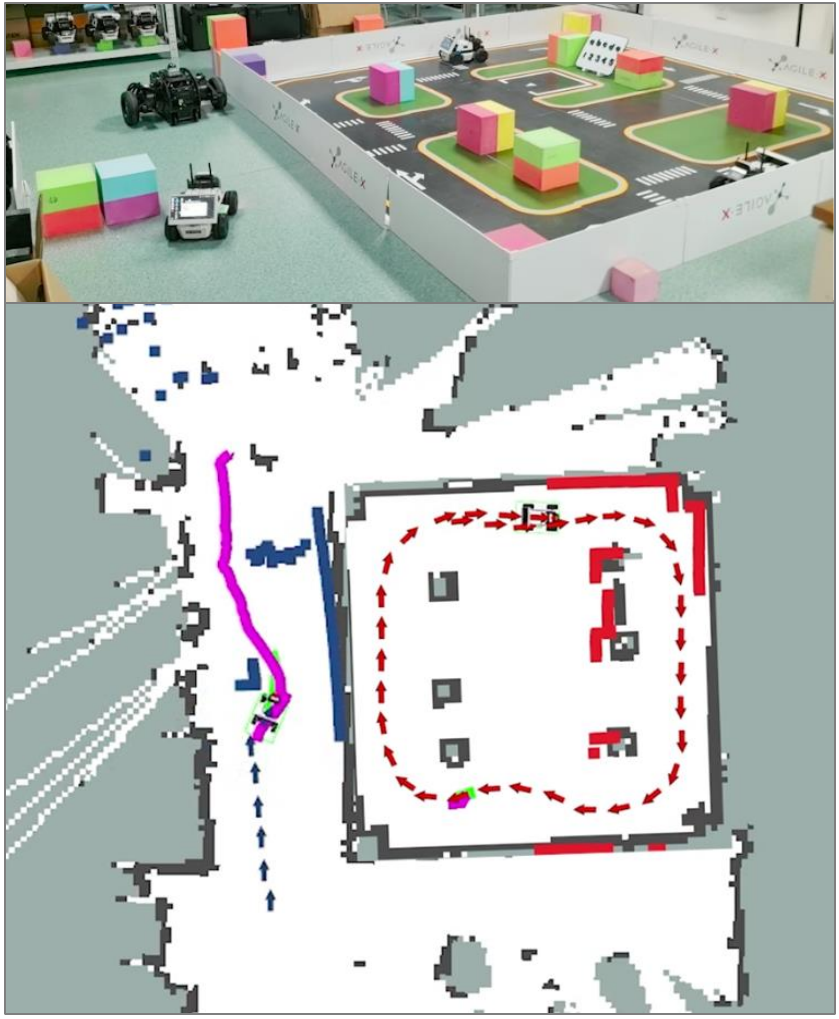}
        \caption{{Robot 2 switches.}}
    \end{subfigure}
    ~
        \begin{subfigure}[t]{0.23\textwidth}
      \includegraphics[width=\textwidth]{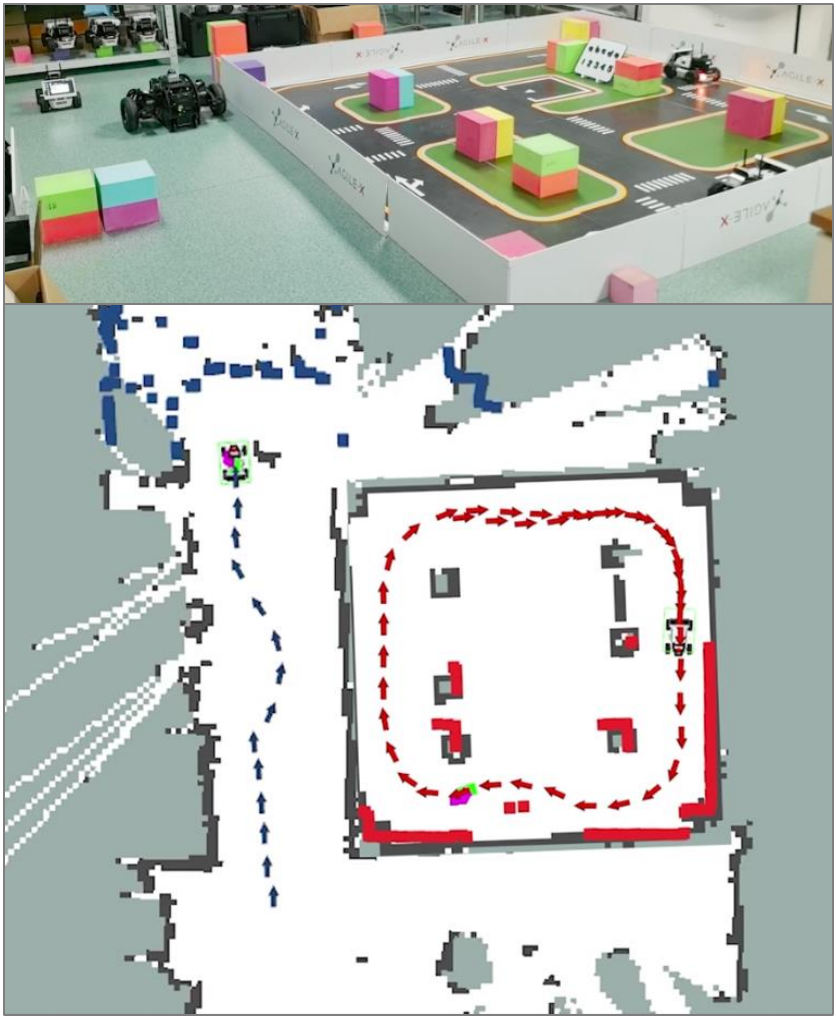}
        \caption{{Task accomplished.}}
    \end{subfigure}
    \caption{{Real-world experiment 2.}
    }
    \label{fig:real_exp_2}
    \vspace{-0.3in}
\end{figure*}

The trajectories of $4$ robots under the proposed EARN are shown at the left and right sides of Fig.~\ref{fig:real_exp}, where the blue and red arrows are generated by local motion planners and the pink paths are generated by edge motion planners.
It can be seen that robots 2 and 4 encounter no obstacles along their target path, and they execute the local motion planner without external assistance until reaching the goal at $t=25.45\,$s and $t=21.45\,$s, respectively.
On the other hand, robots 1 and 3 encounter new obstacles (marked in blue boxes) and leverage planner switching in front of the blue boxes.
Specifically, robot 1 and Diablo server collaboratively accomplish reverse and overtaking actions, so that the robot reaches the goal at $26.00\,$s.
Robot 3 holds its position at the beginning and waits for connection with the server.
At about $t=30\,$s, Diablo moves from the sandbox to the corridor and assists robot 3 for switching to edge motion planning at $t=64\,$s.
With another $13.85\,$s, robot 3 reaches the goal. 
This demonstrates the impact of communication on EARN.

{To validate the effectiveness of switching gain, real experiment 2 is conducted with two LIMO robots and one Diablo server, as depicted in Fig. \ref{fig:real_exp_2}, where robot 1 navigates inside the sandbox, while robot 2 navigates outside the sandbox. As illustrated in Fig. \ref{fig:real_exp_2}a and Fig. \ref{fig:real_exp_2}b, although the robot 2 is closer to the server, its trajectory is planned by the local planner rather than the edge planner. This is because its path is completely blocked by the server, and its switching gain is zero. Consequently, conducting planner switching for this robot leads to a waste of resources. In contrast, planner switching is adopted for robot 1 as shown in Fig. \ref{fig:real_exp_2}b, 
since the local planner gets stuck in front of a box but the edge planner is able to overtake, resulting in enhancement of planning efficiency. 
Furthermore, as illustrated in Fig. \ref{fig:real_exp_2}c, with the movement of server, planner switching also occurs for robot 1 due to the increased switching gain. 
This demonstrates the real-time adaptiveness of EARN.
Finally, the task is successfully completed through collaboration between the server and the robots as depicted in Fig. \ref{fig:real_exp_2}d. Please refer to our video for more details.} 

\section{Conclusion}\label{section6} 

This paper proposed EARN to realize CMP and MPS, thereby opportunistically accelerating the trajectory computations while guaranteeing safety. The new feature of EARN is to optimize robot states and actions under communication and computation resource constraints. CARLA simulations and real-world experiments have shown that EARN reduces the navigation time by $12.1\%$ and $46.7\%$ compared with EDF and PF schemes in indoor and outdoor scenarios, respectively. Furthermore, EARN increases the success rate by $53.1\%$ and $69.0\%$ compared with edge or local computing schemes.

\bibliographystyle{IEEEtran}
\bibliography{reference/Thesis.bib}

\begin{IEEEbiography}
[{\includegraphics[width=1in, height=1.25in, clip, keepaspectratio]{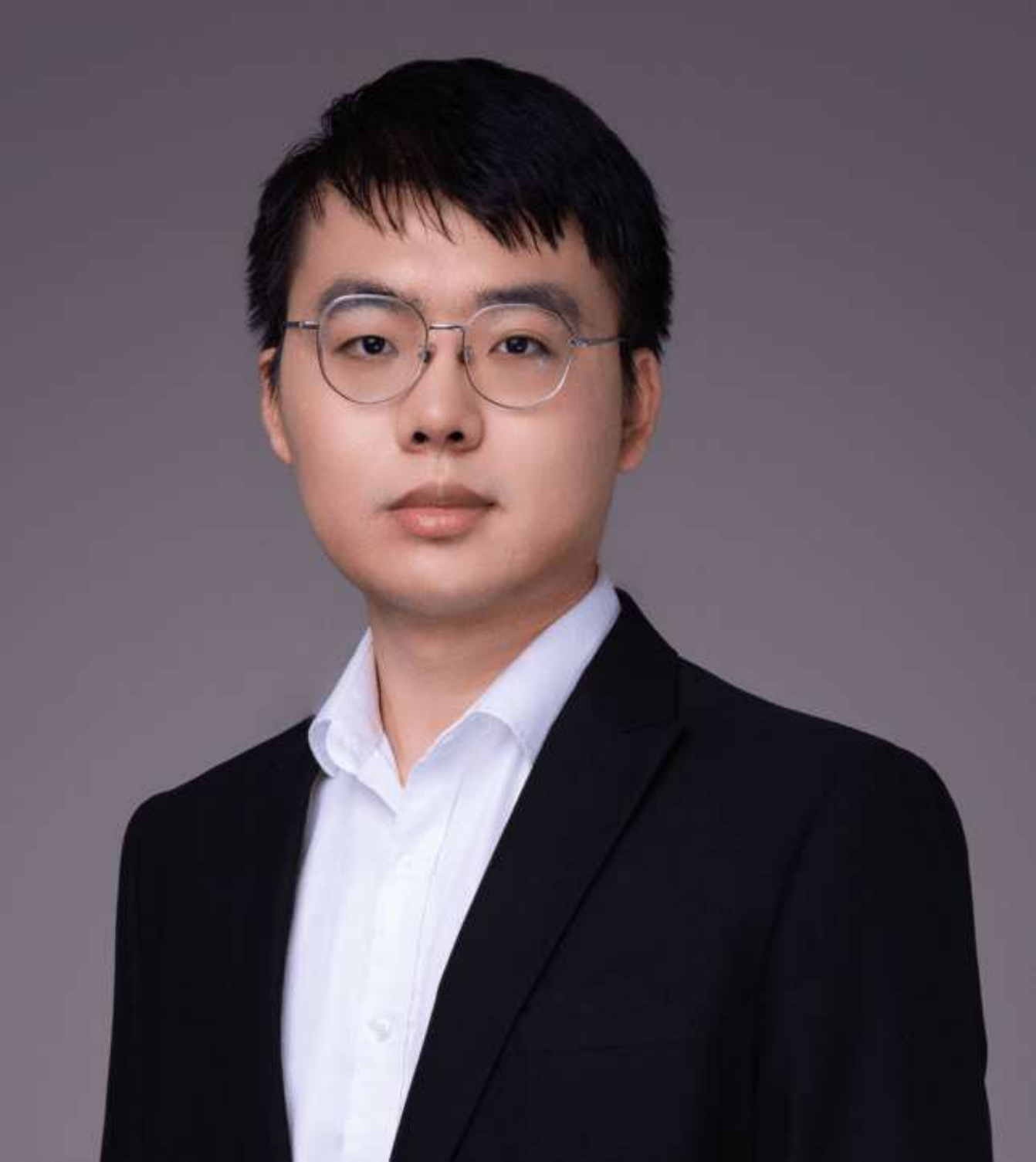}}]{Guoliang Li} (Graduate Student Member, IEEE) received the B.Eng. and M.Phil. degree (\emph{summa cum laude}) from the Southern University of Science and Technology (SUSTech), Shenzhen, China, in 2020 and 2023, respectively. He is currently pursuing the Ph.D. degree in computer science with the University of Macau, Macau, China. His research interests include optimization, motion planning and autonomous systems. He received the National Scholarship in 2022 and the IEEE ICDCS Workshop on SocialMeta Best Student Paper Award in 2023. 
\end{IEEEbiography}	

\begin{IEEEbiography}
[{\includegraphics[width=1in, height=1.25in, clip, keepaspectratio]{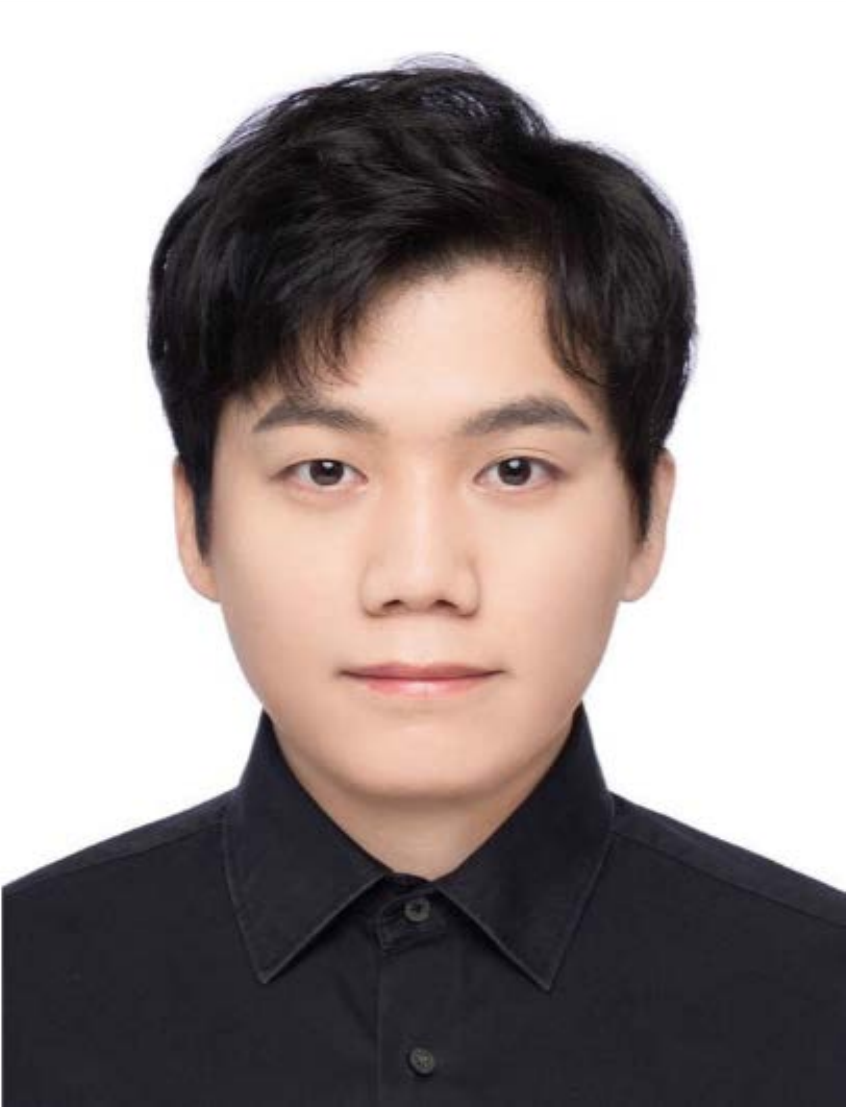}}]{Ruihua Han} (Graduate Student Member, IEEE) received the B.Eng. degree in industrial equipment and control engineering from the Wuhan University of Technology, Wuhan, China and the M.S. degree in microelectronics and solid state electronics from Xiamen University, Xiamen, China. He is currently pursuing the Ph.D. degree in computer science with The University of Hong Kong, Hong Kong. His research interests include optimization, motion planning, reinforcement learning, and autonomous robots.
\end{IEEEbiography}	

\begin{IEEEbiography}
[{\includegraphics[width=1in, height=1.25in, clip, keepaspectratio]{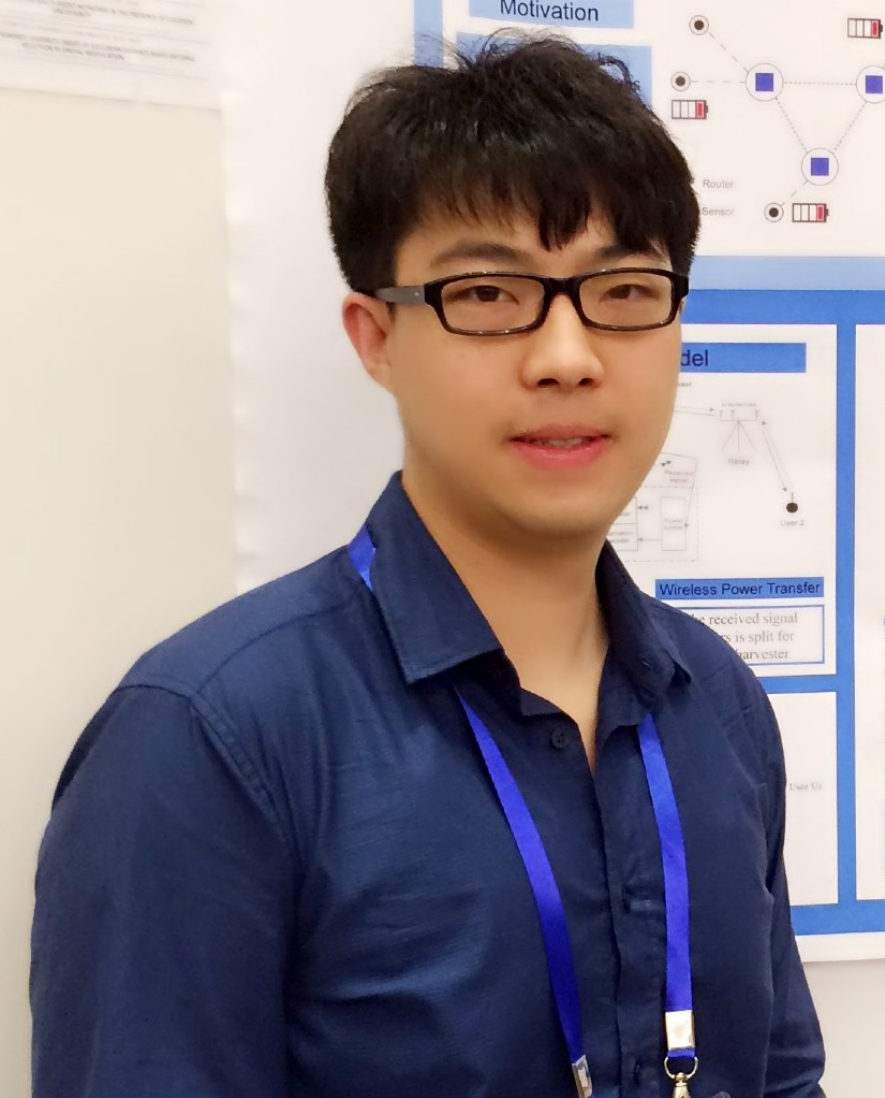}}]{Shuai Wang} (Senior Member, IEEE) received the Ph.D. degree in Electrical and Electronic Engineering from the University of Hong Kong (HKU) in 2018. He is now an Associate Professor with the Shenzhen Institute of Advanced Technology (SIAT), Chinese Academy of Sciences, where he leads the Intelligent Networked Vehicle Systems (INVS) Laboratory. His research interests include autonomous systems and wireless communications. He has published more than 90 journal and conference papers, and received 3 Best Paper Awards from IEEE.
\end{IEEEbiography}

\begin{IEEEbiography}
[{\includegraphics[width=1in, height=1.25in, clip, keepaspectratio]{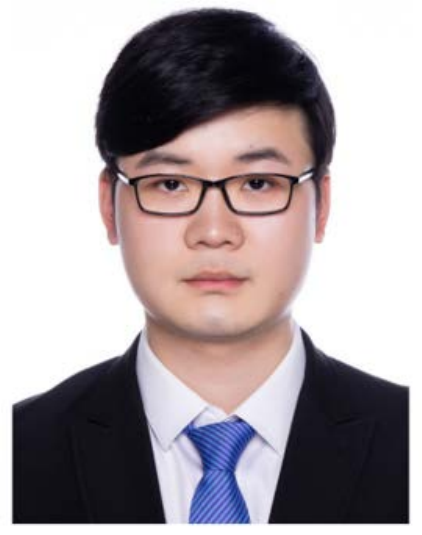}}]{Fei Gao} (Member, IEEE) received the Ph.D. degree in electronic and computer engineering from the Hong Kong University of Science and Technology, Hong Kong, in 2019. He is currently a tenured associate professor at the Department of Control Science and Engineering, Zhejiang University, where he leads the Flying Autonomous Robotics (FAR) group affiliated with the Field Autonomous System and Computing (FAST) Laboratory. His research interests include aerial robots, autonomous navigation, motion planning, optimization, and localization and mapping.
\end{IEEEbiography}

\begin{IEEEbiography}
[{\includegraphics[width=1in, height=1.25in, clip, keepaspectratio]{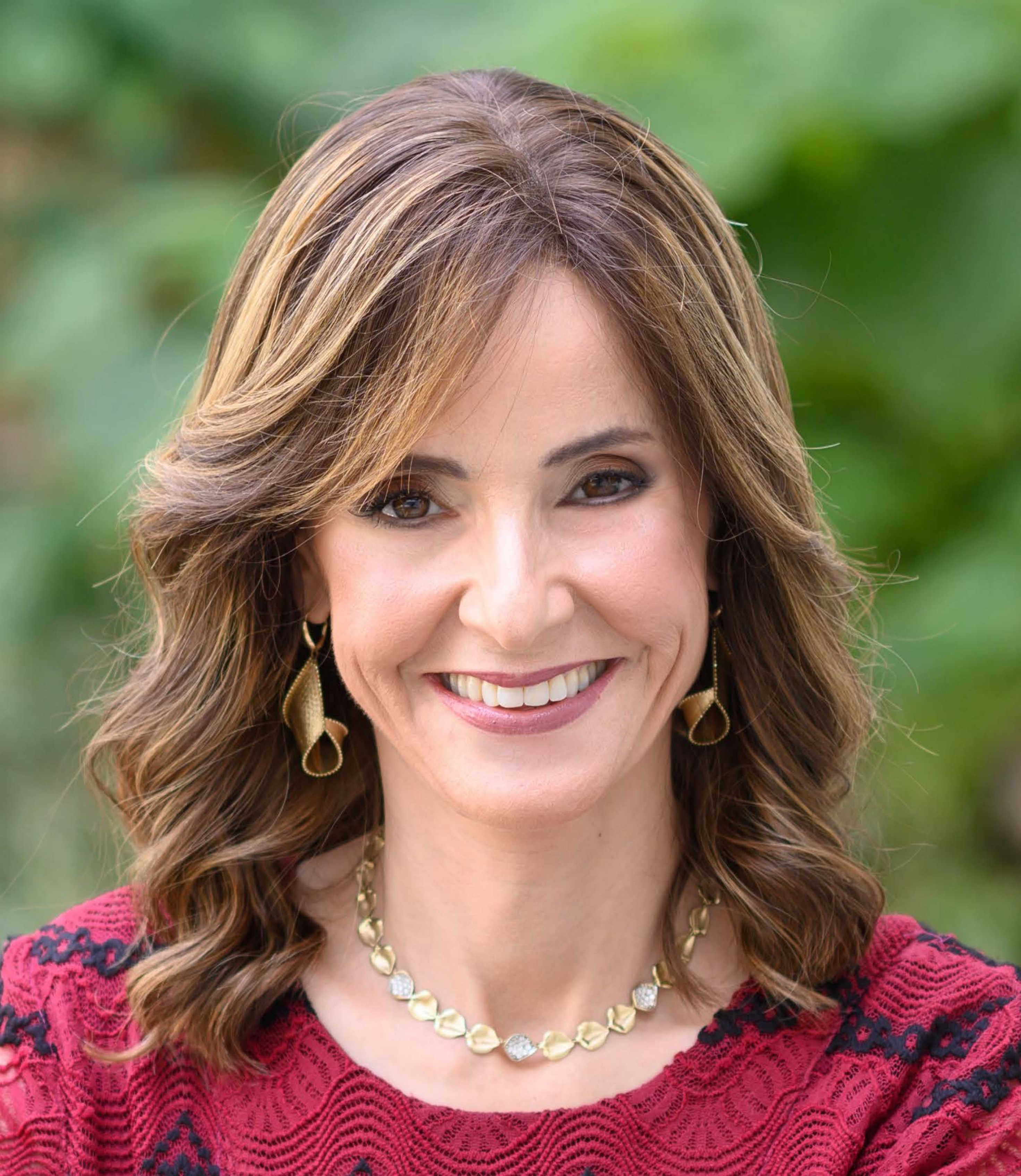}}]{Yonina C. Eldar} (Fellow, IEEE) is a Professor in the Department of Mathematics and Computer Science, Weizmann Institute of Science, Rehovot, Israel where she heads the center for Biomedical Engineering and Signal Processing and holds the Dorothy and Patrick Gorman Professorial Chair. She is also a Visiting Professor at MIT, a Visiting Scientist at the Broad Institute, and an Adjunct Professor at Duke University and was a Visiting Professor at Stanford.  She is a member of the Israel Academy of Sciences and Humanities, an IEEE Fellow and a EURASIP Fellow. She received the B.Sc. degree in physics and the B.Sc. degree in electrical engineering from Tel-Aviv University, and the Ph.D. degree in electrical engineering and computer science from MIT, in 2002. She has received many awards for excellence in research and teaching, including the IEEE Signal Processing Society Technical Achievement Award (2013), the IEEE/AESS Fred Nathanson Memorial Radar Award (2014) and the IEEE Kiyo Tomiyasu Award (2016). She was a Horev Fellow of the Leaders in Science and Technology program at the Technion and an Alon Fellow. She received the Michael Bruno Memorial Award from the Rothschild Foundation, the Weizmann Prize for Exact Sciences, the Wolf Foundation Krill Prize for Excellence in Scientific Research, the Henry Taub Prize for Excellence in Research (twice), the Hershel Rich Innovation Award (three times), and the Award for Women with Distinguished Contributions. She received several best paper awards and best demo awards together with her research students and colleagues, was selected as one of the 50 most influential women in Israel, and was a member of the Israel Committee for Higher Education. She is the Editor in Chief of Foundations and Trends in Signal Processing, a member of several IEEE Technical Committees and Award Committees, and heads the Committee for Promoting Gender Fairness in Higher Education Institutions in Israel.
\end{IEEEbiography}

\begin{IEEEbiography}
[{\includegraphics[width=1in,height=1.25in,clip,keepaspectratio]{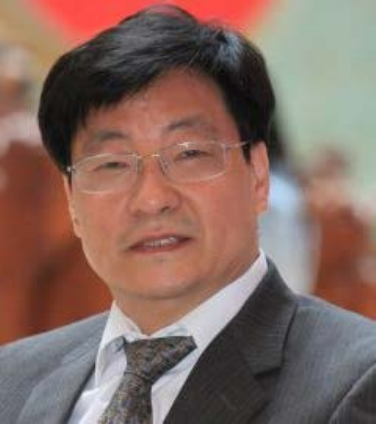}}]{Chengzhong Xu} (Fellow, IEEE) received his Ph.D. degree from the University of Hong Kong in 1993. He is currently a Chair Professor of Computer Science, University of Macau, China. Prior to that, he was in the faculty of Wayne State University and Shenzhen Institutes of Advanced Technology of CAS. His recent research interests are in cloud and distributed computing, intelligent transportation and autonomous driving. He published two research monographs and more than 500 journal and conference papers and received more than 17000 citations. He was a best paper awardee or nominee of conferences, including HPCA’2013, HPDC’2013, ICPP’2015, and SoCC’2021. He was also a co-inventor of more than 150 patents and a co-founder of Shenzhen Institute of Baidou Applied Technology. He serves or served on a number of journal editorial boards, including IEEE TC, IEEE TCC, IEEE TPDS, JPDC, Science China, and ZTE Communication. Dr. Xu was the Chair of IEEE Technical Committee on Distributed Processing from 2015 to 2020. He was a recipient of the Faculty Research Award, Career Development Chair Award, and the President’s Award for Excellence in Teaching of WSU. He was also a recipient of the “Outstanding Oversea Young Scholar” award of NSFC in 2010. Dr. Xu is an IEEE Fellow.
\end{IEEEbiography}

\end{document}